CoilDrop-MRI: Self-supervised physics-guided MRI reconstruction with coil dropout

Tongxi Song[1#], Ziyu Li[2#], Zihan Li[1], Wen Zhong[1], Congyu Liao[3], Yang Yang[3], Hua Guo[1], Wenchuan Wu[2†], Qiyuan Tian[1†,*]

[1]School of Biomedical Engineering, Tsinghua Medicine, Tsinghua University, Beijing, China;

[2]Oxford Centre for Integrative Neuroimaging, FMRIB, Nuffield Department of Clinical Neurosciences, University of Oxford, Oxford, UK;

[3]Department of Radiology & Biomedical Imaging, University of California San Francisco, San Francisco, CA, United States;

[#]Co-first authors. [†]Co-senior authors.

[*]Correspondence to: Qiyuan Tian, Center for Biomedical Imaging Research, Tsinghua University, 30 Shuangqing Road, Haidian District, Beijing, China, 100084 (qiyuantian@tsinghua.edu.cn).

**Highlights**

- CoilDrop-MRI achieves self-supervised MRI reconstruction via random coil partitioning.
- CoilDrop-MRI approaches fully supervised image quality across multiple contrasts and field strengths.
- CoilDrop-MRI enables few-shot learning and robust cross-domain generalization.
- CoilDrop-MRI supports diverse model-based MRI reconstruction frameworks.

**Graphic abstract**

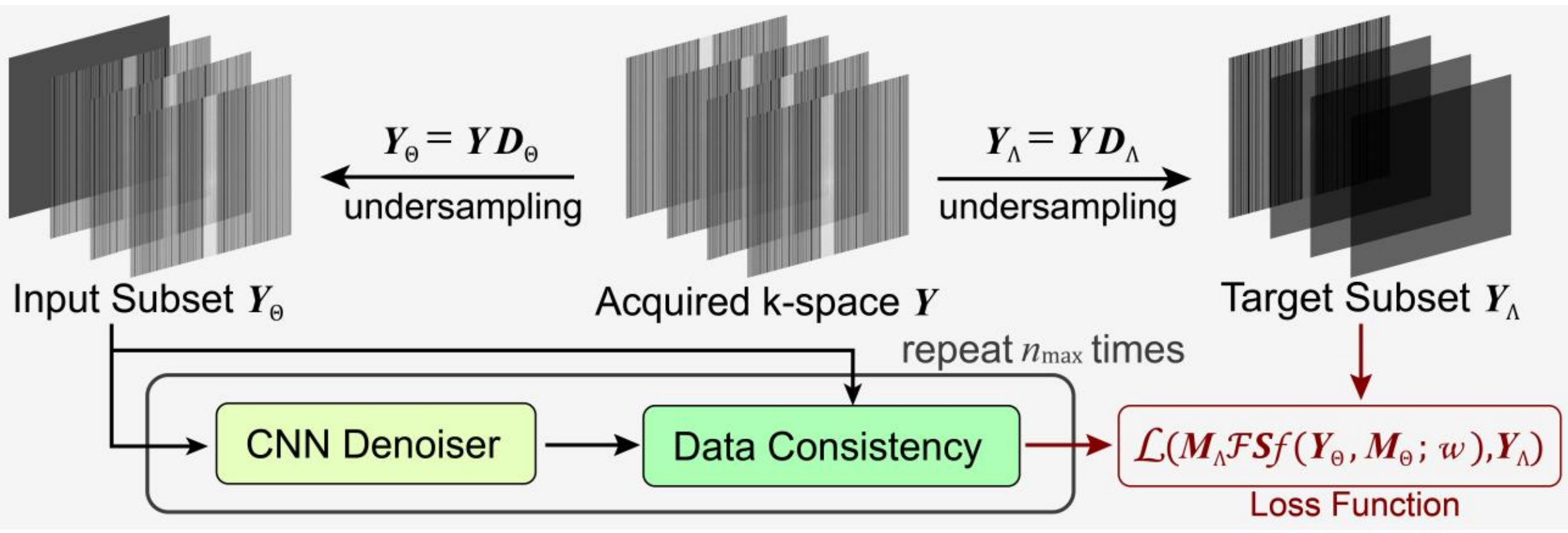

**Abstract**

Self-supervised deep learning-based methods have shown great promise for accelerated magnetic resonance imaging (MRI) reconstruction, achieving high image quality without requiring fully sampled data for training. These methods typically partition the acquired data into two disjoint subsets to construct input–target pairs for optimizing the reconstruction network. However, existing approaches perform this partition exclusively within the spatial frequency (k-space) domain, leaving the coil dimension unexplored. To enforce full exploitation of signal correlation across receiver coils, we propose CoilDrop-MRI, which applies coil-wise dropout to the input and uses the dropped data as training targets in a self-supervised framework. This method is integrated into unrolled architectures in both image-domain (SENSE) and k-space (SPIRiT) formulations. We further demonstrate its versatility by extending CoilDrop-MRI to multi-shot, phase-corrected diffusion MRI (dMRI) reconstruction. CoilDrop-MRI is extensively validated on multi-site, multi-field-strength (0.3T, 0.55T, and 3T), and multi-modality (T1-weighted, T2-weighted, T2-FLAIR, and dMRI) datasets and consistently outperforms state-of-the-art self-supervised methods, achieving quality comparable to supervised reconstruction methods without requiring fully sampled reference training data. Moreover, CoilDrop-MRI exhibits strong data efficiency and robust generalization across imaging conditions, establishing it as a practical and versatile framework for self-supervised parallel MRI reconstruction.

## 1. Introduction

Parallel imaging substantially accelerates magnetic resonance imaging (MRI) by exploiting the spatial sensitivity variations of multiple receiver coils to reconstruct images from undersampled k-space. Existing parallel imaging methods usually fall into two paradigms: image-domain and k-space formulations. Image-domain methods, exemplified by SENSE (Maier et al., 2021; Pruessmann et al., 1999), formulate reconstruction as a linear inverse problem to unfold aliased images using estimated coil sensitivity maps. They are computationally efficient but sensitive to errors in sensitivity estimation. K-space methods, exemplified by GRAPPA (Griswold et al., 2002), synthesize missing k-space data samples across coils with interpolation kernels learned from a fully sampled auto-calibration signal (ACS). They are generally more computationally demanding but more robust to noise amplification and motion-induced sensitivity errors (Hamilton et al., 2017; Lustig & Pauly, 2010). SPIRiT (Lustig & Pauly, 2010) extends GRAPPA by enforcing global self-consistency across multi-coil k-space via iterative optimization. Because each formulation offers distinct advantages, a reconstruction framework applicable to both domains is desirable.

To improve reconstruction performance, prior knowledge has been incorporated through regularization strategies, such as sparsity and low-rank constraints. Sparsity-based methods (Lustig et al., 2007; Yang et al., 2016) assume that images admit sparse representations in specific transform domains (e.g., wavelet) and enforce this property as a regularization constraint to suppress noise and aliasing artifacts. Low-rank-based methods (Haldar, 2013; Haldar & Zhuo, 2016; Li et al., 2025; Mani et al., 2017; Nakarmi et al., 2017) exploit the inherent linear dependencies and structural redundancies within the data and have been used to suppress noise (Haldar, 2013; Haldar & Zhuo, 2016) or correct motion-induced phase inconsistencies (Hu et al., 2019; Li et al., 2025; Mani et al., 2017). However, these hand-crafted priors rely on general assumptions that may not hold for specific datasets or anatomical structures, leading to over-smoothing and residual aliasing artifacts, particularly at high acceleration factors. Moreover, their associated iterative optimization is often computationally expensive and time-consuming.

To overcome these limitations, deep learning methods have emerged as a powerful alternative by introducing data-driven priors. Physics-guided deep learning frameworks, such as Model-Based Deep Learning (MoDL) (Aggarwal et al., 2019) and VarNet (Hammernik et al., 2017), typically replace hand-crafted regularizers with learnable data-driven priors, which can better adapt to the characteristics of specific datasets and reconstruction tasks compared with fixed analytical models. These methods typically unroll iterative optimization algorithms into end-to-end trainable neural networks, alternating between a data consistency (DC) step that enforces the physical forward model and a denoising step that refines the reconstructed image with a convolutional neural network (CNN) (e.g., DnCNN (Zhang et al., 2017)). By combining the known physical forward model with highly expressive data-driven priors, these unrolled networks offer superior noise and artifact suppression, particularly at high acceleration factors. However, a critical bottleneck for supervised model-based methods is their reliance on high-quality, fully sampled reference data, which is often impractical or impossible to acquire in many imaging scenarios.

Self-supervised learning strategies have been developed to address this challenge. A pioneering work in this domain is Self-Supervised learning via Data Undersampling (SSDU) (Yaman et al., 2020). Specifically, SSDU randomly partitions the acquired k-space measurements into two disjoint subsets via dropout masking, predicts the full k-space data from the input subset, and computes the training loss on the target subset to optimize the CNN parameters at each training iteration. Conceptually, SSDU is related to self-supervised image denoising methods (Krull et al., 2019; Lehtinen et al., 2018; Quan et al., 2020) and is particularly closely aligned with Self2Self (Quan et al., 2020), which partitions an image into two subsets and learns to predict one subset from the other. Several variants of SSDU have been proposed to further improve reconstruction performance (Korkmaz et al., 2023; Millard & Chiew, 2024; Yaman, 2022), including improving the denoiser architecture (Korkmaz et al., 2023), optimizing the data partitioning strategies (Yaman, 2022), and combining SSDU with Noisier2Noise. Nevertheless, SSDU and its variants perform the data partitioning only in k-space and do not explicitly exploit inter-coil signal correlations inherent in virtually all MRI acquisitions.

In this work, we propose CoilDrop-MRI, a new self-supervised physics-guided MRI reconstruction framework that incorporates coil dropout to explicitly leverage the coil dimension. Specifically, to exploit the inter-coil signal correlations for self-supervised learning, CoilDrop-MRI reconstructs multi-coil k-space data from a subset of coils, computes the training loss on the remaining coils, and optimizes the network accordingly. Implemented using both SENSE and SPIRiT formulations and evaluated on two low-field datasets and a multi-shot diffusion MRI (dMRI) dataset, CoilDrop-MRI achieves reconstruction quality comparable to that of supervised methods and consistently outperforms Tikhonov-regularized and SSDU-based methods. Moreover, CoilDrop-MRI demonstrates robust generalization across different field strengths and imaging contrasts. Overall, CoilDrop-MRI introduces a new paradigm for self-supervised MRI reconstruction by explicitly exploiting inter-coil redundancy, with the potential to enhance clinical image quality and enable more reliable quantitative brain imaging.

## 2. Theory

### 2.1 Parallel MRI reconstruction

Parallel imaging accelerates MRI acquisition by exploiting the varying spatial sensitivities of receiver coils to recover images from undersampled k-space measurements. A general reconstruction problem can be expressed as:

$$x = \arg\min_{x} \ \mathcal{DC}(x, \boldsymbol{Y}) + \lambda \cdot \mathcal{R}(x)\,, \tag{1}$$

where $x$ denotes the target image to be reconstructed, $\boldsymbol{Y}$ is the acquired undersampled data, $\mathcal{DC}$ represents the data consistency term, and the term $\mathcal{R}(x)$ represents a regularization prior (e.g., sparsity, low-rankness) with $\lambda$ controlling its weight.

Reconstruction algorithms are generally categorized into image-domain and k-space formulations. SENSE (Maier et al., 2021; Pruessmann et al., 1999) is a well-known example of image-domain reconstruction methods, where the unaliased image is recovered by solving the following optimization problem:

$$\arg\min_{x} \|\boldsymbol{M}\mathcal{F}\boldsymbol{S}x - \boldsymbol{Y}\|_2^2 + \lambda \cdot \mathcal{R}(x), \tag{2}$$

where $\boldsymbol{M}$ is the undersampling mask, $\mathcal{F}$ denotes the Fourier transform, and $\boldsymbol{S}$ is the coil sensitivity encoding operator.

In contrast, the k-space reconstruction method SPIRiT (Lustig & Pauly, 2010) exploits local correlations across coils to recover the missing samples in multi-coil k-space, which is achieved by enforcing a self-consistency relation across the entire k-space. To model these relationships, a calibration kernel $\boldsymbol{g}$ is first estimated from the fully sampled ACS data $\widetilde{\boldsymbol{X}}$. The kernel is estimated by minimizing the self-consistency error on the ACS data using a convolutional model:

$$\boldsymbol{g} = \underset{\boldsymbol{g}}{\operatorname{argmin}} \left\|\boldsymbol{g} * \widetilde{\boldsymbol{X}} - \widetilde{\boldsymbol{X}}\right\|_2^2, \tag{3}$$

where $*$ denotes the convolution operation. After calibration, the same self-consistency constraint is applied to the full k-space $\boldsymbol{X}$ to be reconstructed, enforcing $\boldsymbol{g} * \boldsymbol{X} \approx \boldsymbol{X}$. Acquisition consistency is also imposed to ensure agreement between the reconstructed data and the acquired measurements at sampled locations. The SPIRiT reconstruction is then formulated as:

$$\arg\min_{\boldsymbol{X}} \underbrace{\|\boldsymbol{M}\boldsymbol{X}-\boldsymbol{Y}\|_2^2}_{\text{acquisition}} + \underbrace{\mu \cdot \|\boldsymbol{g} * \boldsymbol{X} - \boldsymbol{X}\|_2^2}_{\text{calibration}} + \underbrace{\lambda \cdot \mathcal{R}(\boldsymbol{X})}_{\text{regularization}}, \quad (4)$$

where $\mu$ controls the trade-off between the acquisition consistency term and the calibration consistency term. The final image $x$ can be obtained from the reconstructed k-space data $\boldsymbol{X}$ via inverse Fourier transform followed by coil combination.

The optimization problems in Eqs. (2) and (4) are typically solved using iterative algorithms, such as ISTA (Guerquin-Kern et al., 2011) and ADMM (Boyd et al., 2011), which can be computationally demanding and time-consuming.

**2.2 Model-based deep learning (MoDL) reconstruction**

MoDL reconstruction combines CNN-based learned regularization with the MRI physics-based forward model, replacing conventional hand-crafted priors:

$$x = \arg\min_{x} \mathcal{DC}(x, \boldsymbol{Y}) + \lambda \cdot \|x - \mathcal{D}_w(x)\|_2^2, \quad (5)$$

where $\mathcal{DC}$ denotes the data consistency term, and $\mathcal{D}_w$ is a CNN denoiser with learnable weights $w$. The regularization term $\mathcal{R}(x) = \|x - \mathcal{D}_w(x)\|_2^2$ acts as a data-driven image prior by penalizing the difference between the reconstructed image $x$ and its denoised version $\mathcal{D}_w(x)$, thereby suppressing noise and aliasing artifacts.

To solve this optimization problem, an alternating minimization scheme is adopted, and its iterations are unrolled into an $N$-stage neural network cascade. To prevent the model size from scaling with the number of iterations, the weights $w$ of the denoiser are shared across all stages. The reconstruction proceeds by alternating between a CNN-based denoising step and a DC step. Specifically, at the $n$-th iteration, the updates are given by:

$$z_n = \mathcal{D}_w(x_n), \tag{6}$$

$$x_{n+1} = \arg\min_x \mathcal{DC}(x, \boldsymbol{Y}) + \lambda \cdot \|x - z_n\|_2^2. \tag{7}$$

The implementation of the DC step in Eq. (7) depends on the choice of forward model. While the original MoDL implementation (Aggarwal et al., 2019) employed a SENSE-based formulation, this study extends it to both SENSE-based (Fig. 1a) and SPIRiT-based (Fig. 1b) formulations, with the corresponding update rules for the DC layers given by:

$$x_{n+1} = \begin{cases} (\boldsymbol{S}^H \mathcal{F}^{-1} \boldsymbol{M} \mathcal{F} \boldsymbol{S} + \lambda \boldsymbol{I})^{-1} (\boldsymbol{S}^H \mathcal{F}^{-1} \boldsymbol{M} \boldsymbol{Y} + \lambda z_n); \text{SENSE}, \\ \boldsymbol{S}^H (\mathcal{F}^{-1} \boldsymbol{M} \mathcal{F} + \mu (\boldsymbol{G} - \boldsymbol{I})^H (\boldsymbol{G} - \boldsymbol{I}) + \lambda \boldsymbol{I})^{-1} (\mathcal{F}^{-1} \boldsymbol{M} \boldsymbol{Y} + \lambda (\boldsymbol{S} z_n)); \text{ SPIRiT}, \end{cases} \tag{8}$$

where $\boldsymbol{G}$ is the inverse Fourier transform of the calibration kernel $\boldsymbol{g}$, and $\boldsymbol{I}$ denotes the identity mapping. $\boldsymbol{S}$ maps the denoised single-coil image to the multi-coil domain, whereas $\boldsymbol{S}^H$ maps the multi-coil SPIRiT reconstruction back to a single-coil image for subsequent CNN denoising.

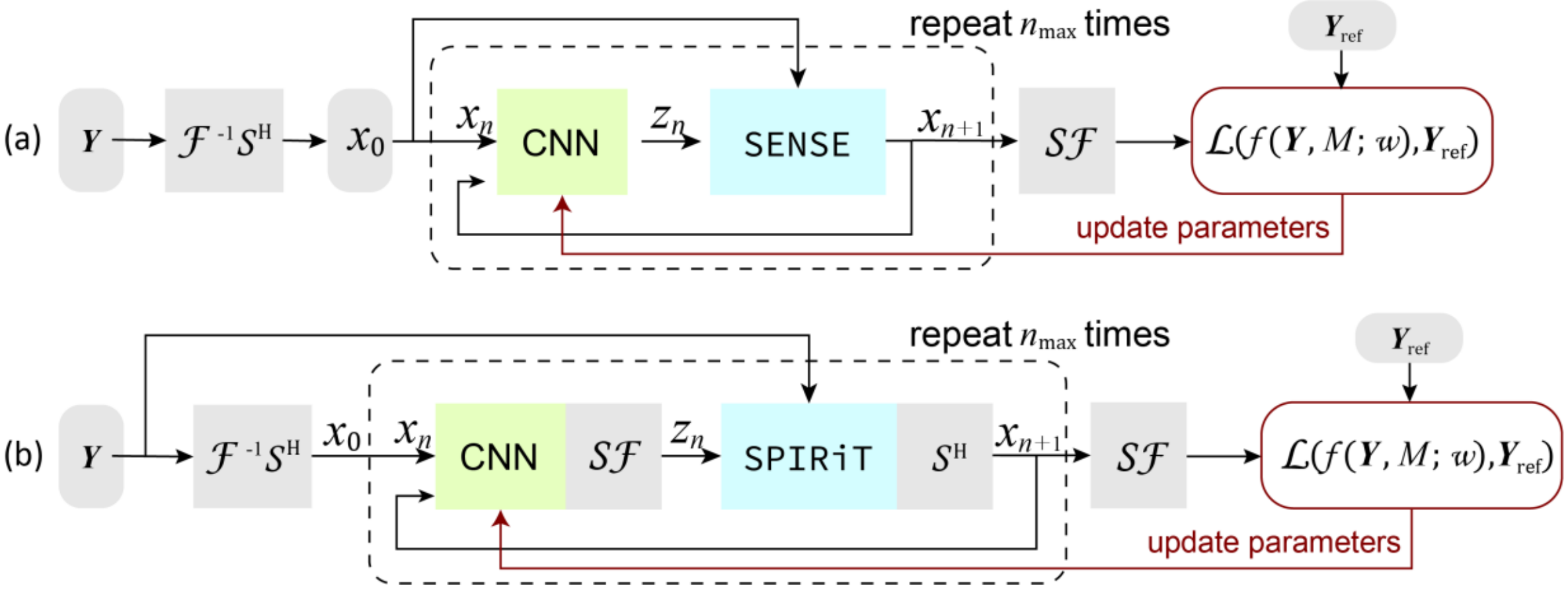

**Figure 1. Model-based reconstruction.** The model-based reconstruction problem is solved using an unrolled network, which alternately performs convolutional neural network (CNN)-based denoising and data consistency (DC) steps. The DC step is implemented in either the image-domain using SENSE (a) or in k-space using SPIRiT (b).

### 2.3 Self-supervised learning

While MoDL reconstruction demonstrates superior performance over conventional methods, it requires fully sampled reference data for supervised training. Self-supervised learning has emerged as a promising alternative for training physics-guided networks without reference data. A prominent example is SSDU, which partitions k-space indices $\Omega$ into two disjoint subsets: input subset $\Theta$ and target subset $\Lambda$, using stochastic binary masks $\boldsymbol{D}_\Theta$ and $\boldsymbol{D}_\Lambda$ which satisfy $\boldsymbol{D}_\Theta + \boldsymbol{D}_\Lambda = \mathbf{1}$, where $\mathbf{1}$ denotes an all-ones tensor of the same size as the k-space data. The data and corresponding undersampling masks from the input ($\boldsymbol{Y}_\Theta,\ \boldsymbol{M}_\Theta$) and the target ($\boldsymbol{Y}_\Lambda,\ \boldsymbol{M}_\Lambda$) subsets are generated via element-wise multiplication:

$$\begin{cases} \boldsymbol{Y}_\Lambda = \boldsymbol{D}_\Lambda \boldsymbol{Y}, \boldsymbol{M}_\Lambda = \boldsymbol{D}_\Lambda \boldsymbol{M} \\ \boldsymbol{Y}_\Theta = \boldsymbol{D}_\Theta \boldsymbol{Y}, \boldsymbol{M}_\Theta = \boldsymbol{D}_\Theta \boldsymbol{M} \end{cases}. \tag{9}$$

During training, the network $f(\cdot; w)$ takes $\boldsymbol{Y}_\Theta$ and $\boldsymbol{M}_\Theta$ as inputs and reconstructs an image, which is then mapped back to k-space via the forward operator $\mathcal{F}\boldsymbol{S}$ to compute the training loss against $\boldsymbol{Y}_\Lambda$ (Fig. 2a):

$$\min_w \mathcal{L}(\boldsymbol{M}_\Lambda \mathcal{F} \boldsymbol{S} f(\boldsymbol{Y}_\Theta, \boldsymbol{M}_\Theta; w), \boldsymbol{Y}_\Lambda). \tag{10}$$

Multi-coil k-space data possesses inherent information redundancy arising from inter-coil correlations and structured k-space representations. The efficacy of exploiting this redundancy depends strongly on the partitioning scheme employed. To identify the optimal partitioning strategy for robust reconstruction, a progression of partitioning strategies is systematically investigated (Fig. 2b), from k-space partitioning (SSDU) to joint k-space and coil partitioning (Coil-Incoherent SSDU, CI-SSDU), and finally to coil-only partitioning (CoilDrop).

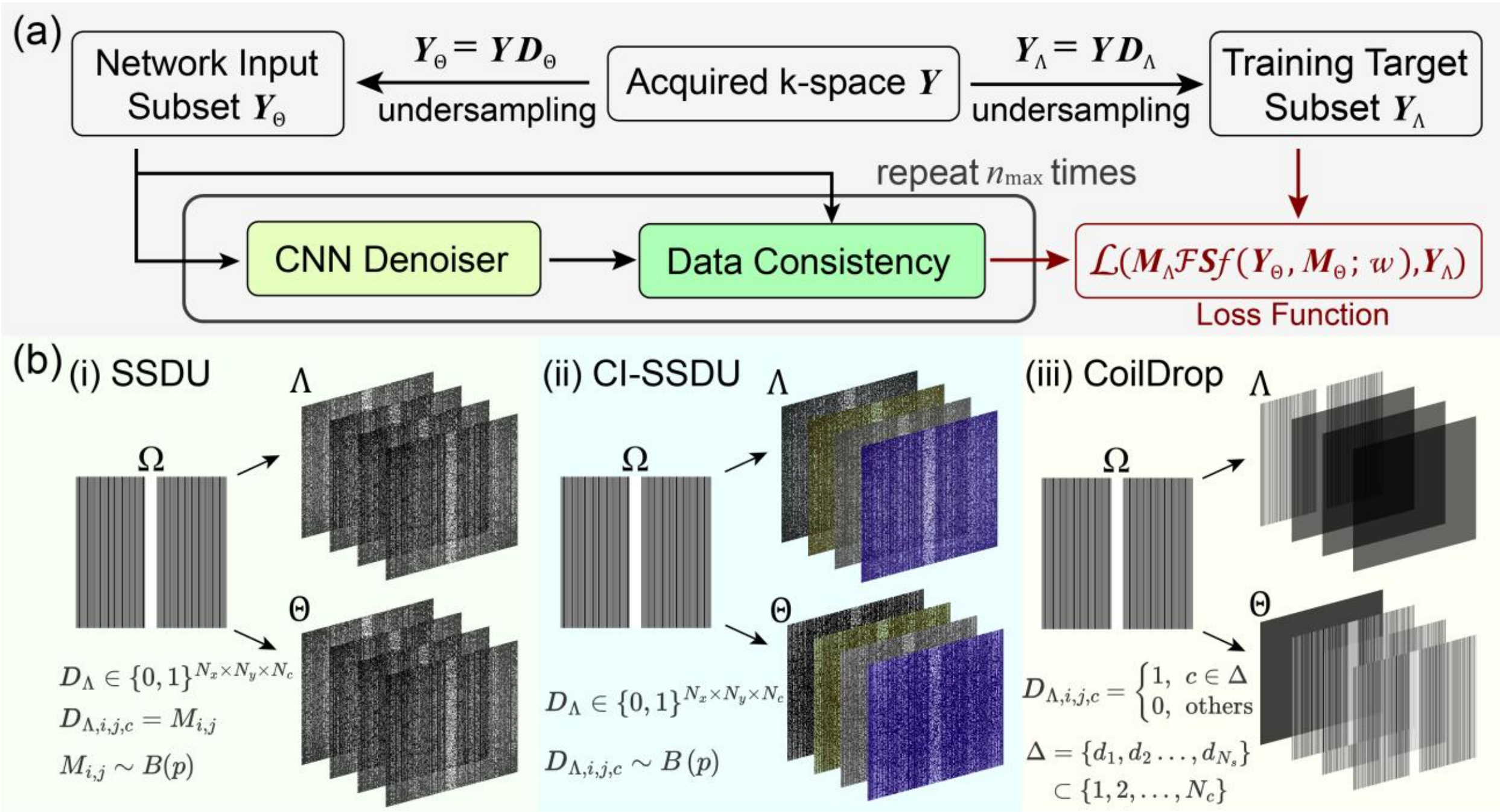

**Figure 2. Self-supervised learning via data partitioning.** (a) The undersampled, multi-coil data are divided into two disjoint subsets (Θ and Λ). The model receives data from subset Θ as input and is trained by minimizing the prediction error in subset Λ. (b) Three data partitioning strategies are illustrated, including (i) SSDU partitions data in the $k_x - k_y$ plane using a shared random mask across all coils; (ii) Coil-Incoherent SSDU (CI-SSDU) partitions data in the $k_x - k_y$ plane using independent random masks for each coil (indicated by the varying colors of the masks across the coil dimension); (iii) CoilDrop-MRI randomly partitions the data along the coil dimension.

### 2.3.1 SSDU

The original SSDU approach employs a consistent partitioning mask across all coil channels (Fig. 2b, i), with $\boldsymbol{D}_\Lambda \in \{0,1\}^{N_x \times N_y \times N_c}$ defined as:

$$\boldsymbol{D}_{\Lambda;i,j,c} = M_{i,j}, \text{ where } M_{i,j} \sim B(p)\,, \tag{11}$$

where $B(p)$ denotes a Bernoulli distribution with probability $p$.

Under this formulation, partitioning is purely in k-space, with all coils sharing the same mask. When a specific k-space location $(i, j)$ is assigned to the target set Λ, measurements at that location are withheld across all coils. Notably, since all coils are partitioned identically, this strategy does not explicitly exploit the inter-coil redundancy inherent in multi-coil MRI. This observation motivates the exploitation of the coil dimension for improved self-supervised reconstruction, so as to more fully leverage the rich signal correlations across receiver channels.

### 2.3.2 CI-SSDU

As an intermediate step, we introduce and evaluate a joint k-space and coil partitioning strategy, Coil-Incoherent SSDU (CI-SSDU), which bridges pure k-space and coil-only approaches (Fig. 2b, ii). Specifically, the mask is generated independently across coils as $\boldsymbol{D}_{\Lambda;i,j,c} \sim B(p)$, allowing each coil to have a distinct sampling pattern. Consequently, a given k-space location may be dropped and assigned to the target subset for some coils while remaining in the input subset for others, thereby encouraging the network to exploit inter-coil signal correlations.

### 2.3.3 CoilDrop

CoilDrop is proposed to fully exploit the information redundancy in the coil dimension (Fig. 2b, iii), partitioning the data entirely along the coil dimension by randomly selecting a subset of coils:

$$\Delta = \{d_1, d_2 \dots, d_{N_s}\} \subset \{1,2, \dots, N_c\}, \tag{12}$$

as the training target subset $\Lambda$, with the remaining coils forming the input subset $\Theta$. The mask is defined as:

$$\boldsymbol{D}_{\Lambda;i,j,c} = \begin{cases} 1 & c \in \Delta, \\ 0 & \text{otherwise.} \end{cases} \tag{13}$$

Under this formulation, the network predicts the data of the target coils $\Delta$ using only the complementary input coils $\{1, \dots, N_c\} \setminus \Delta$, enabling explicit exploitation of inter-coil information redundancy.

## 3. Methods and experiments

### 3.1 Data collection and preprocessing

#### 3.1.1 M4Raw 0.3T data

The primary validation was performed using the publicly available preprocessed low-field M4Raw dataset (Lyu et al., 2023) to demonstrate CoilDrop-MRI's efficacy under low-SNR conditions. The data were acquired on a 0.3T whole-body MRI system (Oper-0.3, Ningbo Xingaoyi Medical Instruments Co., Ltd., Ningbo, China) equipped with a 4-channel head coil. The dataset includes T1-weighted (T1w), T2-weighted (T2w), and T2-FLAIR image data acquired from healthy volunteers with in-plane resolution = 1.2 × 1.2 mm² and slice thickness = 5 mm. Each scan was repeated 2 or 3 times without using parallel imaging or partial Fourier acquisition. Acquisition and processing parameters were described in detail in the original paper (Lyu et al., 2023). Multi-repetition data within each scan were co-registered to the first repetition using the "flirt" function in the FMRIB Software Library (FSL) (Smith et al., 2004) with 6 degrees of freedom and then averaged to obtain a high-SNR reference. The network input was restricted solely to the first repetition of each scan during training, validation, and evaluation. For the experiments, 80 subjects were randomly selected from the dataset, including 50 for training, 10 for validation (only used for the supervised method), and 20 for evaluation.

#### 3.1.2 UCSF 0.55T data

To assess the generalization capability of CoilDrop-MRI across different scanners and field strengths, T2-FLAIR Turbo Spin Echo (TSE) data were collected from 5 subjects on a 0.55T Siemens MAGNETOM Free.Max scanner at the University of California, San Francisco (UCSF) using either 10- or 12-channel head coils. Written informed consent was obtained from all participants in accordance with local institutional ethics approval. Data were acquired without using parallel imaging or partial Fourier acquisition, with the following sequence parameters: TR = 7000 ms, TE = 80 ms, inversion time (TI) = 2200 ms, FOV = 240 × 196 mm², in-plane resolution = 0.9 × 0.9 mm², and slice thickness = 5 mm. Noise scans without RF excitation were acquired for noise pre-whitening of the multi-coil k-space data. To ensure a consistent number of channels across all subjects, the pre-whitened data were subsequently compressed to 10 virtual channels (Zhang et al., 2012). For all subjects, the image matrix size was standardized to 256 × 224.

### 3.1.3 THU dMRI data

To demonstrate the effectiveness of CoilDrop-MRI on multi-shot dMRI reconstruction, data were collected from one subject on a 3T Siemens MAGNETOM Prisma MRI scanner using a 52-channel head coil at Tsinghua University (THU), with written informed consent obtained in accordance with local institutional ethics approval. Two b = 0 volumes were acquired at the beginning of the sequence, followed by 32 diffusion-weighted image (DWI) volumes at b = 1000 s/mm$^2$ using a 4-shot interleaved echo-planar imaging (EPI) sequence with the following sequence parameters: TR = 3000 ms, TE = 63 ms, FOV = 212 × 212 mm$^2$, in-plane resolution = 1.2 × 1.2 mm$^2$, slice thickness = 5 mm, partial Fourier = 3/4, and no parallel imaging. The raw k-space data were pre-whitened using a noise scan. To obtain a high-SNR reference, the acquisition was repeated three times, and the resulting data were co-registered to the first repetition using FSL's "flirt" function and then averaged.

### 3.1.4 Data undersampling simulation

To simulate accelerated MRI acquisitions, retrospective undersampling was performed on the M4Raw dataset using three configurations: variable-density (VD) sampling with acceleration factors of 2 (R = 2) and 3 (R = 3), along with equispaced sampling with R = 3 (Supplementary Fig. S1). Given that M4Raw was acquired with only four coil channels, R = 3 represents a relatively high acceleration factor for this dataset. For VD undersampling, the ACS region was retained in the reconstruction input. For equispaced undersampling, which was used to simulate EPI acquisition, the ACS region was excluded from the reconstruction input and used exclusively for sensitivity-map estimation using ESPIRiT (Uecker et al., 2014) and SPIRiT kernel calibration. The ESPIRiT calibration was implemented using the "ecalib" function in the Berkeley Advanced Reconstruction Toolbox (Uecker et al., 2015).

For out-of-distribution evaluation, the UCSF T2-FLAIR data were retrospectively undersampled using VD with R = 3. For THU dMRI data, 4-fold equispaced undersampling was applied only to the DWI volumes. The b = 0 volumes remained fully sampled, with the central 8% of k-space data used for sensitivity-map estimation.

### 3.2 Experimental setup

Five experiments were conducted to systematically evaluate CoilDrop-MRI across five aspects: reconstruction accuracy, data efficiency, cross-modality generalization, out-of-distribution robustness, and dMRI applicability.

#### 3.2.1 Reconstruction accuracy

CoilDrop-MRI was compared against SSDU, Robust SSDU (Millard & Chiew, 2024), and CI-SSDU on the M4Raw dataset in terms of reconstructed image quality, evaluated across multiple undersampling patterns (Section 3.1.4), imaging contrasts (T1w, T2w, and T2-FLAIR), and reconstruction formulations (SENSE, SPIRiT). Supervised MoDL and Tikhonov-regularized reconstruction were also included as representative supervised learning-based and conventional optimization-based baselines, respectively. The comparison methods and their implementations are briefly described as follows.

**SSDU:** The original SSDU framework (Yaman et al., 2020) was implemented as a self-supervised baseline for model-based MRI reconstruction. Specifically, the data partitioning for SSDU was performed using a 50:50 k-space split into input and target subsets.

**Robust SSDU:** The recently proposed Robust SSDU (Millard & Chiew, 2024) method was also compared, which integrates the Noisier2Noise principle (Millard & Chiew, 2023) into the SSDU framework to achieve simultaneous self-supervised reconstruction and denoising. Similar to the original SSDU, it utilized a 50:50 k-space split. Following its Noisier2Noise principle, synthetic complex Gaussian noise $\tilde{n} \sim \mathcal{CN}(0, \alpha^2\sigma_n^2)$ was injected into the input subset during training, with the hyperparameter α set to 0.05.

**CI-SSDU:** To evaluate the effect of incoherent masking, CI-SSDU was implemented following the joint k-space and coil partitioning strategy described in Section 2.3.2 with a 50:50 multi-coil k-space split. This method applies independent stochastic masking for each coil channel instead of employing an identical mask across all coils as in the original SSDU.

**Tikhonov-Regularized Reconstruction:** Tikhonov-regularized SENSE and SPIRiT were employed as non-learning reconstruction baselines. This method solves the inverse problem by minimizing a linear least-squares objective with an $l_2$-norm regularizer:

$$x = \arg\min_{x} \mathcal{DC}(x, \boldsymbol{Y}) + \lambda \cdot \|x\|_2^2 \tag{14}$$

The $\mathcal{DC}$ term can be implemented in either the SENSE-based or SPIRiT-based formulation, depending on the choice of forward model. The optimization problem was solved using the conjugate gradient (CG) algorithm.

**Supervised MoDL:** The supervised MoDL method, trained with fully sampled data as supervision, was included to serve as an empirical performance upper bound. The training followed the setup in the original study (Aggarwal et al., 2019), minimizing the pixel-wise mean squared error (MSE) loss between the reconstruction and the fully sampled reference.

### 3.2.2 Data efficiency

Few-shot learning experiments were conducted under R = 2 VD undersampling with the SPIRiT formulation. For each modality (T1w, T2w, and T2-FLAIR), the number of training subjects was systematically reduced from 50 to 40, 30, 20, 10, 5, 2, and 1, and all trained models were evaluated on the fixed, held-out 20-subject test set.

### 3.2.3 Cross-modality generalization

To assess the cross-modality generalization of CoilDrop-MRI, the model was trained exclusively on a single modality and directly applied to the other two unseen modalities under R = 2 VD undersampling with the SPIRiT formulation. Furthermore, a mixed-modality training strategy was investigated by constructing a heterogeneous training set of 50 volumes (17 T1w, 17 T2w, and 16 T2-FLAIR volumes). The trained model was then evaluated on the test sets of all three modalities.

### 3.2.4 Out-of-distribution robustness

To evaluate CoilDrop-MRI's generalization to out-of-distribution datasets, experiments were conducted on the 5-subject UCSF T2-FLAIR dataset under R = 3 VD undersampling using both the SENSE and SPIRiT formulations. First, the supervised MoDL, SSDU, and proposed CoilDrop-MRI models pre-trained on the M4Raw T2-FLAIR data under the same undersampling configuration were directly applied to the unseen UCSF data. Subsequently, the pre-trained SSDU and CoilDrop-MRI models were fine-tuned on the UCSF data by minimizing the corresponding loss functions. The fine-tuning configurations were kept identical to those used for the initial M4Raw training, and the fine-tuned models were evaluated to assess performance gains relative to direct application without fine-tuning.

### 3.2.5 Diffusion MRI applicability

To demonstrate CoilDrop's applicability to dMRI, two tasks were evaluated: (1) single-shot reconstruction with 4-fold acceleration, assessing the proposed method's ability to enable fast high-SNR dMRI; and (2) phase-corrected multi-shot joint reconstruction, assessing its capacity to accommodate shot-to-shot phase variations in the forward model. CoilDrop-MRI models for dMRI reconstruction were implemented in SENSE formulation and trained from scratch, independent of the models trained on the M4Raw and UCSF datasets.

For single-shot dMRI reconstruction, all settings followed those used on the M4Raw dataset, except that R = 4 equispaced undersampling was employed. For multi-shot dMRI reconstruction, a MUSE-type two-stage self-supervised framework was adopted (Supplementary Fig. S2). In the first stage, Unrolled Network I was trained using all four shots; during each training update, one shot was randomly selected and its coil data were partitioned according to the CoilDrop strategy for self-supervised learning. Subsequently, the trained model was applied to each shot to reconstruct four complex-valued images. Phase maps $\boldsymbol{P} = \{P_1, P_2, \dots, P_s\}$ were extracted from these reconstructed images as $P_j = x_j/|x_j|$ and then smoothed using a 32 × 32 k-space Hamming window. In the second stage, the estimated phase maps and the multi-shot data were fed into the forward model of Unrolled Network II under the CoilDrop-MRI framework, yielding the final high-fidelity DWIs. The results were corrected for eddy currents and head motion using the "eddy" (Andersson & Sotiropoulos, 2016) function in FSL, and DTI metrics were subsequently computed using the "dtifit" function in FSL.

### 3.3 Implementation details

All learning-based reconstruction models were implemented in PyTorch based on the open-source fastMRI repository (Zbontar et al., 2018) and trained on an NVIDIA A800 GPU. The MoDL architecture was adopted as the common backbone and unrolled for $N = 10$ iterations, with a 10-layer 2D CNN used as the regularization unit (Supplementary Fig. S3). In each unrolled iteration, the data consistency subproblem was solved using the conjugate gradient algorithm with 30 iterations.

All network parameters were optimized using the Adam optimizer with a batch size of 8 and an initial learning rate of $5 \times 10^{-4}$. The learning rate was dynamically adjusted using a ReduceLROnPlateau scheduler with a reduction factor of 0.5 and a patience of 5 epochs, with the training loss used as the monitored metric. Each network was trained for 50 epochs, requiring approximately 8 hours for SENSE-based implementations and 12 hours for SPIRiT-based implementations. For evaluation, self-supervised methods utilized the final epoch's weights, whereas supervised methods used the checkpoint with the lowest validation loss. To ensure reproducibility, the random seed was fixed to 42 and deterministic operations were enforced.

The reconstruction hyperparameters were set as learnable parameters and initialized consistently across deep learning models, while they were fixed for the conventional baselines. For both supervised and self-supervised models, the regularization coefficient $\lambda$ in Eq. (5) was treated as a learnable parameter initialized to 0.05. For SPIRiT-based reconstruction, the calibration consistency coefficient $\mu$ was also learnable and initialized to 1.0. For the Tikhonov-regularized baselines, these hyperparameters were fixed, with $\lambda = 0.05$ for SENSE, and $\lambda = 0.05,\ \mu = 1.0$ for SPIRiT.

For CoilDrop-MRI, self-supervised data partitioning was performed dynamically and exclusively along the coil dimension during training. In each training iteration, 3/4 (rounded up) of the receiver coils were randomly assigned to the input subset $\Theta$, while the remaining coils were assigned to the target subset $\Lambda$.

### 3.4 Evaluation metrics

Quantitative metrics including mean absolute error (MAE), peak signal-to-noise ratio (PSNR), and structural similarity index (SSIM) were calculated to measure the similarity between the reconstructed image and the high-SNR reference. For the M4Raw and multi-shot dMRI datasets, the high-SNR references were obtained by averaging co-registered multi-repetition acquisitions. For the 0.55T UCSF dataset, the high-SNR reference was obtained by applying BM4D (Maggioni et al., 2012) denoising to the fully sampled reconstructions, assuming a noise standard deviation of 0.02. For the DTI metrics in the multi-shot dMRI experiments, the MAEs of fractional anisotropy (FA), mean diffusivity (MD), and axial diffusivity (AD) were computed between the metrics derived from the reconstructed images and those obtained from the high-SNR reference.

## 4. Results

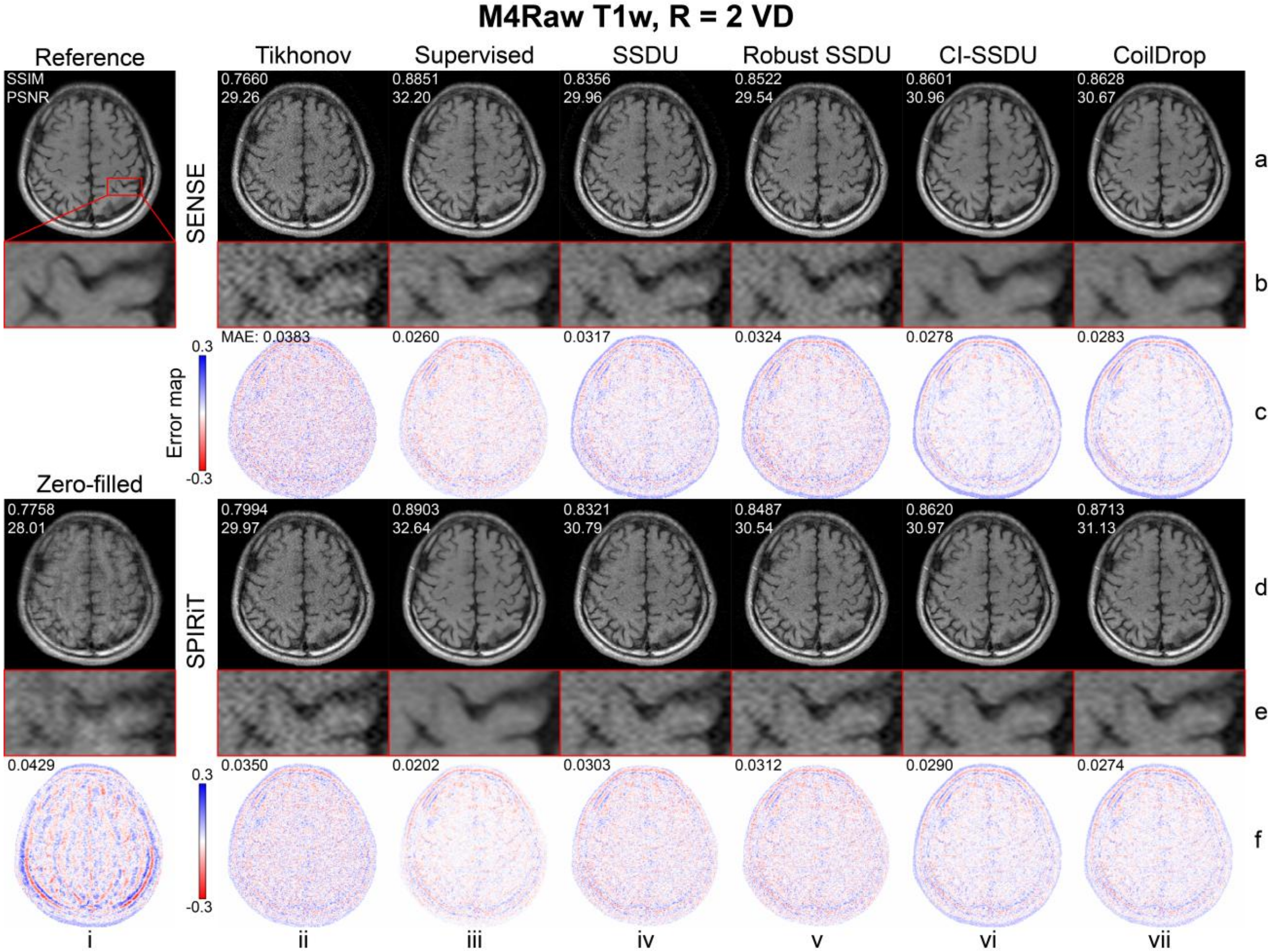

**Figure 3. Comparison of reconstruction methods with variable density (VD) sampling at 2× acceleration (R = 2) for T1-weighted (T1w) images.** Representative reconstructed images, zoomed-in regions, and corresponding error maps are shown for R = 2 VD T1w images from one test subject. Both SENSE- and SPIRiT-based implementations are compared for the following: multi-repetition-averaged high-SNR reference (top left), zero-filled reconstruction (bottom left), Tikhonov-regularized reconstruction, supervised MoDL, and four self-supervised strategies: SSDU, Robust SSDU, Coil-Incoherent SSDU (CI-SSDU), and CoilDrop. Quantitative metrics, including structural similarity index (SSIM), peak signal-to-noise ratio (PSNR), and mean absolute error (MAE) with the high-SNR reference are calculated to quantify reconstruction fidelity.

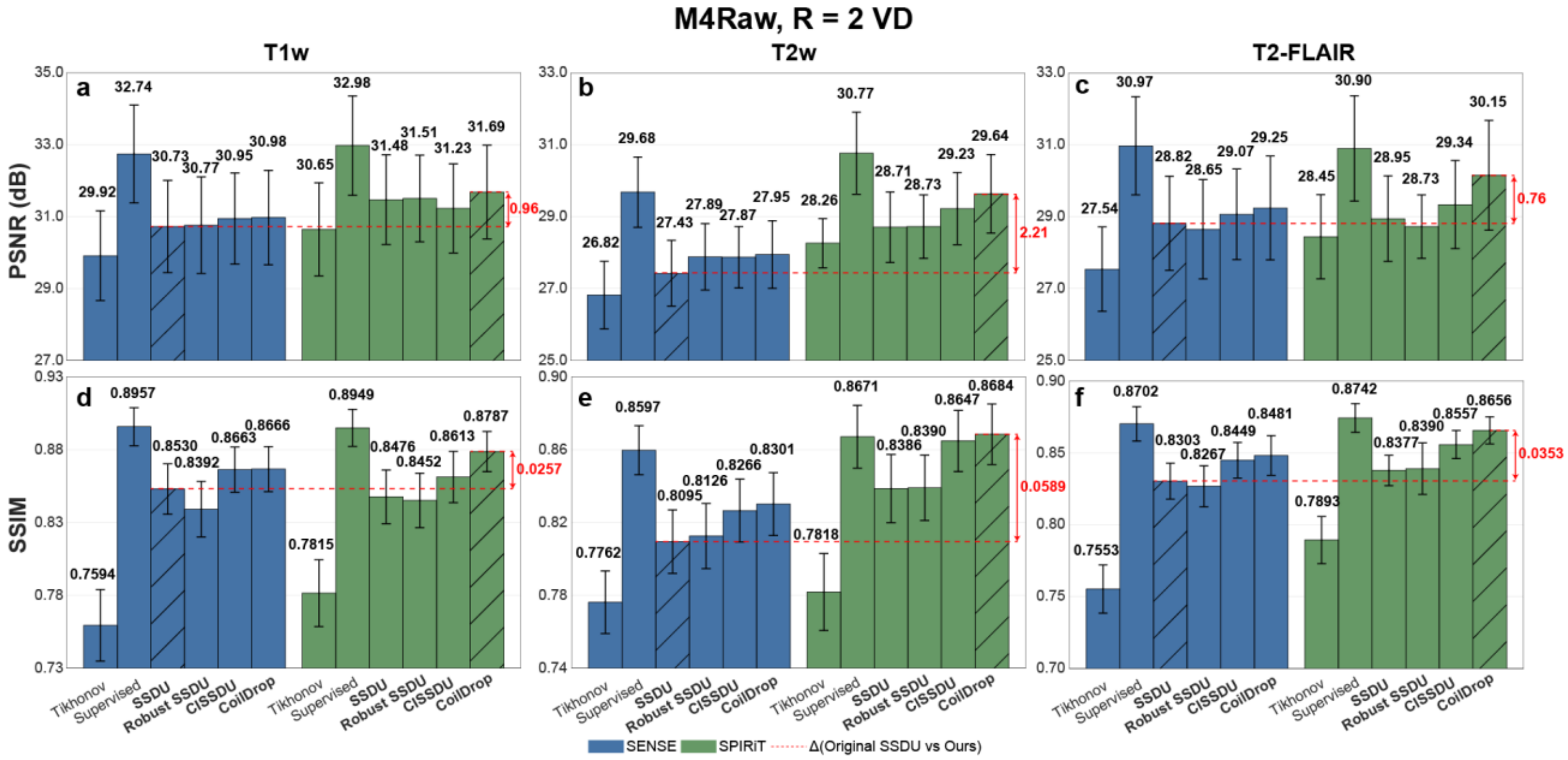

**Figure 4. Quantitative comparison of reconstruction methods under variable density (VD) sampling at 2× acceleration (R = 2).** Bar charts illustrate the average (a-c, top row) peak signal-to-noise ratio (PSNR) and (d-f, bottom row) structural similarity index (SSIM) across the test subjects for (a, d) T1-weighted (T1w), (b, e) T2-weighted (T2w), and (c, f) T2-FLAIR images under R = 2 VD reconstructions. Blue and green bars represent reconstruction frameworks employing SENSE- and SPIRiT-based data consistency, respectively. The evaluated methods include the conventional Tikhonov-regularized baseline, the supervised MoDL, and four self-supervised strategies (marked in **bold**): SSDU, Robust SSDU, Coil-Incoherent SSDU (CI-SSDU), and the proposed CoilDrop (highlighted with diagonal hatches). Red dashed lines and text indicate the quantitative performance gains of the proposed CoilDrop-MRI over the original SSDU method. Error bars denote standard deviations.

### 4.1 Reconstruction accuracy

Figure 3 compares the reconstruction performance of different methods on T1w images with 2-fold VD undersampling. Conventional Tikhonov-regularized reconstructions suffer from noticeable residual noise (Fig. 3, ii), which is effectively suppressed by the supervised MoDL (Fig. 3, iii). Existing self-supervised methods, including SSDU and Robust SSDU, also improve reconstruction quality relative to the Tikhonov baseline (Fig. 3, iv, v), but still show elevated noise compared to the high-SNR reference and MoDL. In contrast, the proposed CI-SSDU and CoilDrop-MRI produce high-quality reconstructions with improved noise suppression and well-preserved image textures, approaching the performance of the supervised MoDL (Fig. 3, vi, vii). Our implementations work robustly under both the SENSE and SPIRiT formulations, with SPIRiT-based reconstructions showing slightly cleaner results, presumably because of the improved noise conditioning from k-space reconstruction (Lustig & Pauly, 2010). Representative sagittal reconstructions are provided in Supplementary Fig. S4.

The quantitative evaluations agree with visual observations (Fig. 4). Tikhonov-regularized reconstructions yield relatively low PSNR and SSIM, which are improved substantially by supervised MoDL, providing an upper performance bound given access to fully sampled training data. Self-supervised methods achieve intermediate performance, with CI-SSDU and CoilDrop-MRI consistently outperforming existing SSDU-based approaches. CoilDrop-MRI further provides modest but consistent gains over CI-SSDU, highlighting the advantage of exploiting only the coil dimension over joint k-space and coil partitioning. Notably, the SPIRiT-based implementations achieve higher PSNR and SSIM than their SENSE-based counterparts. In particular, our SPIRiT-based CoilDrop-MRI improves PSNR over the original SSDU by 0.96 dB, 2.21 dB, and 0.76 dB for T1w, T2w, and T2-FLAIR images, respectively.

For the more challenging case of R = 3 equispaced undersampling, which represents a relatively high acceleration factor given that M4Raw was acquired with only four coil channels, similar trends are observed (Figs. 5, 6). All reconstruction methods suffer from degradation due to the higher acceleration factor, with more pronounced noise and aliasing artifacts in the Tikhonov and conventional SSDU reconstructions. Robust SSDU even underperforms the original SSDU in this setting. In contrast, the proposed CI-SSDU and CoilDrop-MRI maintain robust performance, with visibly improved noise suppression and structural fidelity. Notably, CoilDrop-MRI shows a more pronounced advantage in this regime, yielding clearer image details and higher quantitative metrics compared to existing self-supervised approaches. These results further highlight the effectiveness of exploiting the coil dimension, particularly under more aggressive undersampling conditions. The reconstruction results for R = 3 VD (Supplementary Figs. S5, S6) also follow similar trends.

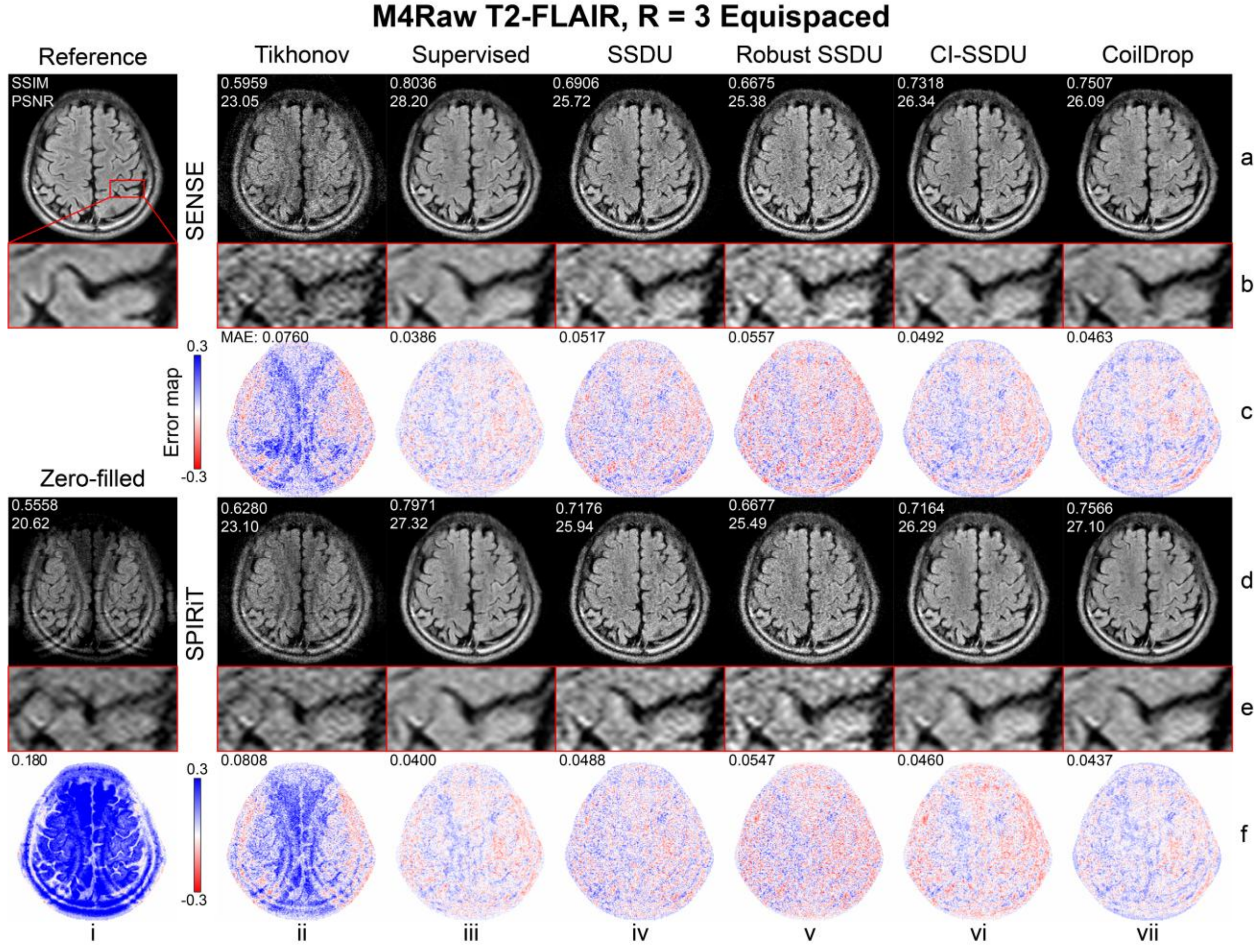

**Figure 5. Comparison of reconstruction methods under equispaced sampling at 3× acceleration (R = 3) for T2-FLAIR images.** Representative reconstructed images, zoomed-in regions, and corresponding error maps are shown for T2-FLAIR images from one test subject. The display layout, evaluated methods, and annotated quantitative metrics are identical to those detailed in Figure 3.

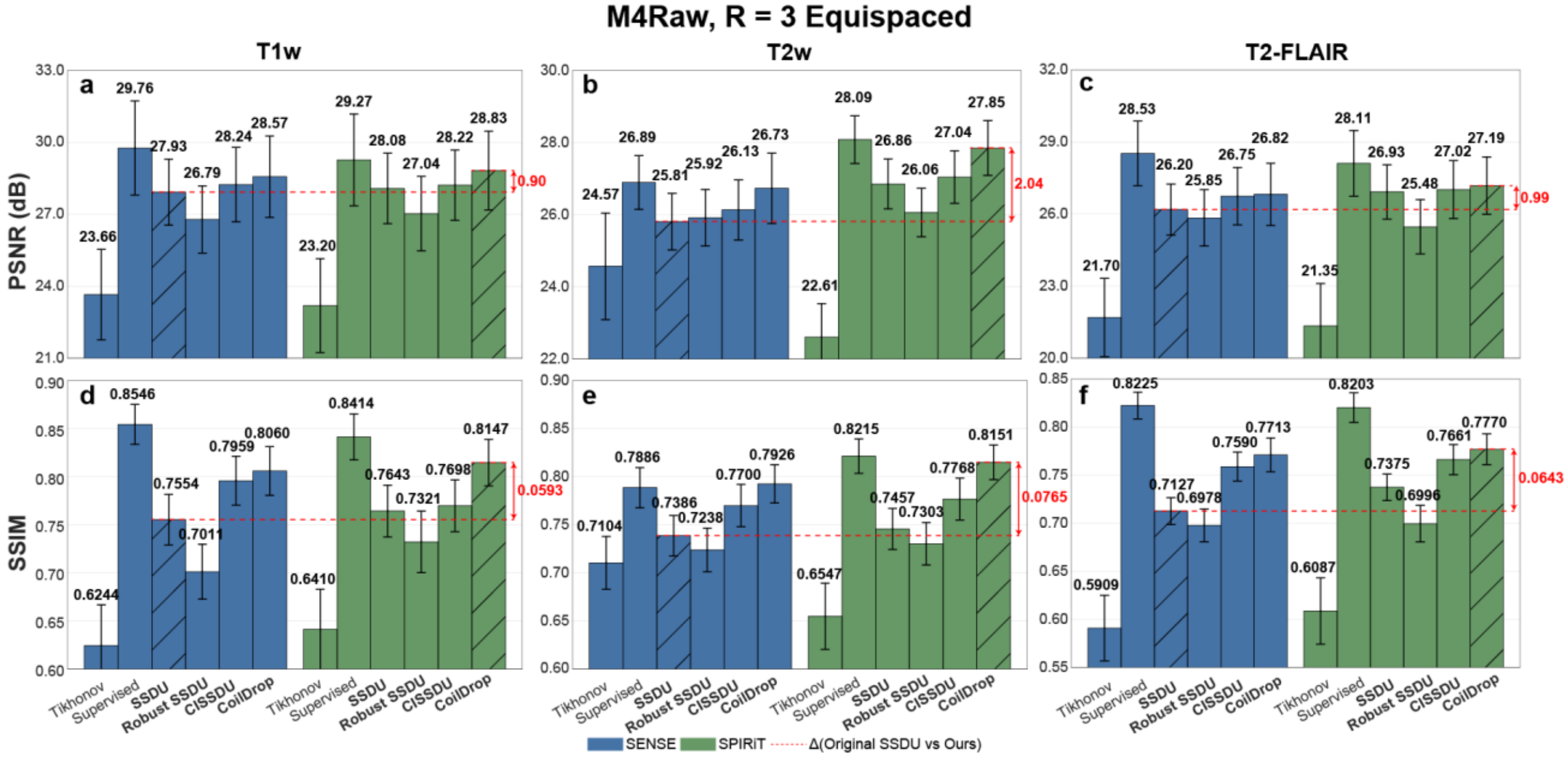

**Figure 6. Quantitative comparison of reconstruction methods under equispaced sampling at 3× acceleration (R = 3).** Bar charts illustrate the average peak signal-to-noise ratio (PSNR) and structural similarity index (SSIM) across the test subjects for T1-weighted (T1w), T2-weighted (T2w), and T2-FLAIR images. The chart layout, evaluated methods, and graphical conventions are identical to those detailed in Figure 4.

### 4.2 Data efficiency

CoilDrop-MRI demonstrates robust performance even with reduced training subjects (Fig. 7). Under R = 2 VD sampling, reconstruction quality improves rapidly as the number of training subjects increases from 1 to 10, with average PSNR rising from 28.9 dB to 31.2 dB. Beyond this point, performance begins to plateau, showing only modest gains to 31.7 dB as the number of training subjects increases to 50. This rapid early improvement indicates that CoilDrop-MRI is highly data-efficient, achieving high-quality reconstructions with only a small number of training subjects. Notably, CoilDrop-MRI surpasses the Tikhonov baseline when trained on as few as five subjects. Similar trends are consistently observed across T1w, T2w, and T2-FLAIR contrasts.

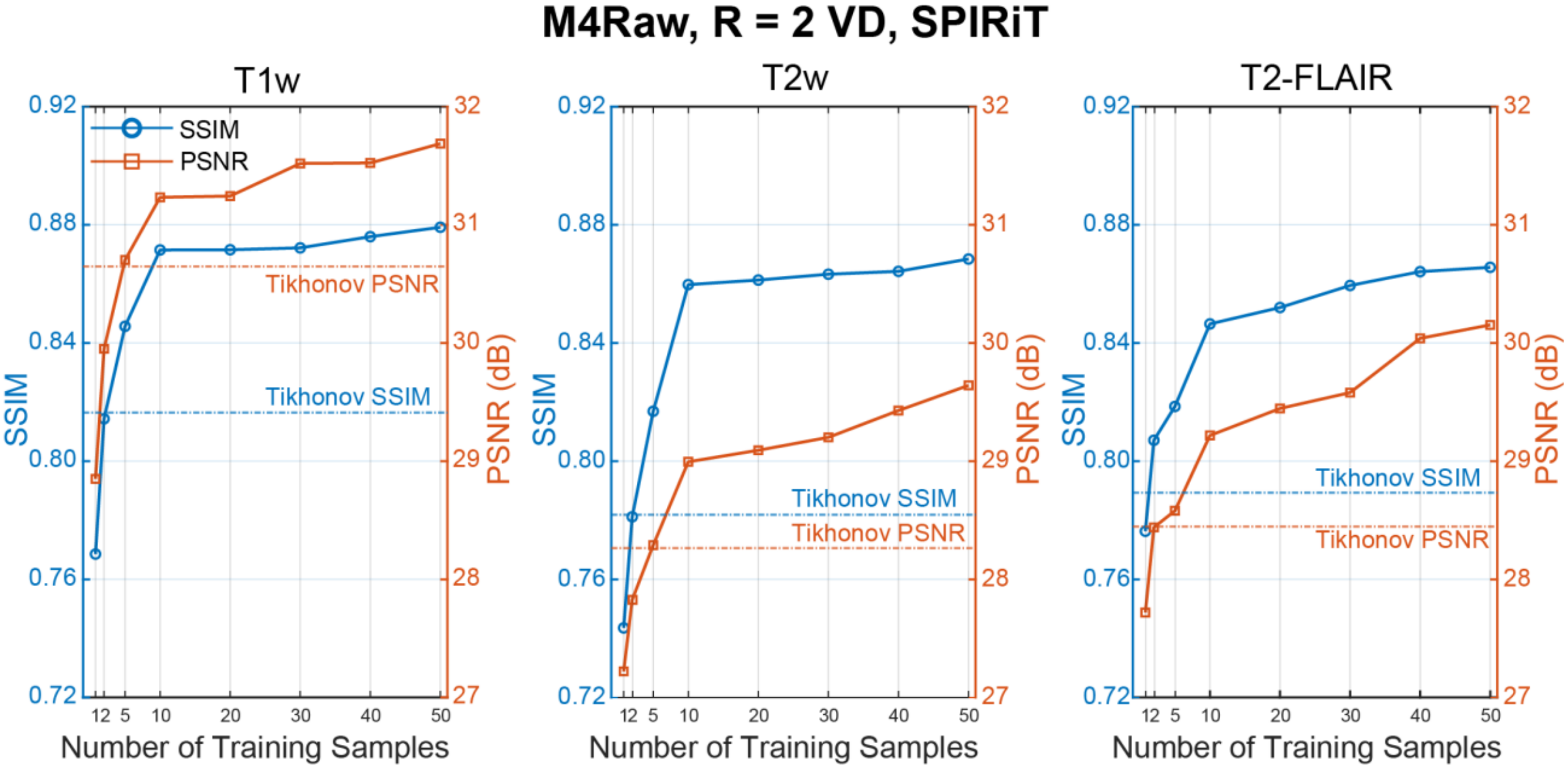

**Figure 7. CoilDrop-MRI few-shot learning performance.** CoilDrop-MRI performance is plotted as a function of the number of training subjects for T1w, T2w, and T2-FLAIR images, quantified by SSIM (blue curves, left axes) and PSNR (red curves, right axes) relative to the high-SNR reference. All experiments were conducted using the SPIRiT-based formulation of CoilDrop-MRI at 2× acceleration with variable density sampling. The horizontal dashed lines represent the baseline performance achieved by Tikhonov-regularized reconstruction.

### 4.3 Cross-modality generalization

CoilDrop-MRI also exhibits strong cross-modality generalization (Table 1). As expected, reconstruction performance is highest when the training and testing modalities are matched (diagonal entries). Models trained on a single modality maintain robust performance during zero-shot inference on unseen contrasts. The PSNR degradation remains no greater than 0.70 dB across all cases, indicating reliable generalization. Notably, all cross-modality results substantially exceed the Tikhonov baseline (PSNR = 30.65, 28.26, and 28.45 dB for T1w, T2w, and T2-FLAIR, respectively). The mixed-modality training strategy further improves performance, achieving the highest overall average across all modalities. Compared to modality-specific training (diagonal entries), the mixed model incurs only a negligible PSNR reduction of no more than 0.23 dB.

**M4Raw, R = 2 VD, SPIRiT, Cross-modality Performance (PSNR)**

| Test Modality \ Train Modality | T1w | T2w | T2-FLAIR | Mixed |
|---|---|---|---|---|
| T1w | **31.69±1.30** | 31.21±1.37 | 30.99±1.14 | 31.46±1.35 |
| T2w | 29.03±1.09 | **29.64±1.09** | 29.24±0.81 | 29.56±1.09 |
| T2-FLAIR | 29.46±1.37 | 29.63±1.37 | **30.15±1.53** | 29.96±1.55 |
| Average | 30.06 | 30.16 | 30.13 | **30.33** |

**M4Raw, R = 2 VD, SPIRiT, Cross-modality Performance (SSIM)**

| Test Modality \ Train Modality | T1w | T2w | T2-FLAIR | Mixed |
|---|---|---|---|---|
| T1w | **0.8803±0.0140** | 0.8653±0.0149 | 0.8624±0.0150 | 0.8711±0.0157 |
| T2w | 0.8533±0.0171 | **0.8684±0.0167** | 0.8548±0.0186 | 0.8613±0.0166 |
| T2-FLAIR | 0.8592±0.0107 | 0.8567±0.0101 | **0.8656±0.0094** | 0.8649±0.0088 |
| Average | 0.8643 | 0.8635 | 0.8609 | **0.8657** |

**Table 1. CoilDrop-MRI cross-modality generalization performance.** Group-level (mean ± standard deviation across 10 test subjects) peak signal-to-noise ratio (PSNR, top) and structural similarity index (SSIM, bottom) between the CoilDrop-MRI reconstructed images and the high-SNR reference are shown for models trained on single or mixed modalities (columns) and evaluated on different test contrasts (rows). **Bold** values indicate the best performance in each row.

**UCSF FLAIR, R = 3 VD, Out-of-Distribution Generalization Performance**

| | | Zero-Shot | | Self-Supervised Finetuning | |
|---|---|---|---|---|---|
| Reconstruction Formulation | Method | PSNR (dB) | SSIM | PSNR (dB) | SSIM |
| SENSE | Supervised | 30.30±1.81 | 0.9063±0.0210 | Not Applicable | Not Applicable |
| | SSDU | 29.38±1.48 | 0.8662±0.0094 | 32.03±1.84 | 0.8636±0.0379 |
| | CoilDrop | 29.41±1.73 | 0.8798±0.0125 | 32.05±1.99 | 0.8941±0.0416 |
| SPIRiT | Supervised | 30.29±2.42 | 0.8865±0.0498 | Not Applicable | Not Applicable |
| | SSDU | 31.07±2.14 | 0.8768±0.0577 | 31.70±1.94 | 0.8864±0.0554 |
| | CoilDrop | 31.09±1.72 | 0.9008±0.0130 | 32.65±1.35 | 0.9081±0.0105 |

**Table 2. Quantitative evaluation of CoilDrop-MRI generalization to out-of-distribution UCSF 0.55T T2-FLAIR data.** Results are shown for direct zero-shot inference of supervised MoDL, SSDU, and CoilDrop, as well as for target-domain self-supervised fine-tuning of SSDU and CoilDrop, each implemented in both SENSE- and SPIRiT-based formulations. Reconstruction quality was evaluated using peak signal-to-noise ratio (PSNR) and structural similarity index (SSIM), reported as mean ± standard deviation across five test subjects. Both metrics were computed against the corresponding BM4D-denoised fully sampled reference images.

### 4.4 Out-of-distribution robustness

CoilDrop-MRI also generalizes well to out-of-distribution 0.55T data acquired from the UCSF dataset with a different scanner and protocol ($N = 5$, R = 3 VD, Fig. 8). Direct zero-shot inference of models pre-trained on 0.3T data successfully recovers general anatomical structures for the supervised MoDL and the self-supervised methods (SSDU and CoilDrop-MRI), but exhibits noticeable blurring due to differences in scanners and protocol setups (Fig. 8b, c). Despite this domain shift, CoilDrop-MRI consistently maintains a modest quantitative and visual advantage over SSDU across both SENSE and SPIRiT formulations. Self-supervised fine-tuning on the target 0.55T subjects effectively mitigates domain-shift residual artifacts without requiring any fully sampled reference data. Improvements are noticeable in SENSE- and SPIRiT-based SSDU reconstructions. While self-supervised fine-tuning noticeably improves SSDU reconstructions under both formulations, residual blurring persists, particularly with SPIRiT (Fig. 8d). CoilDrop-MRI further mitigates these artifacts, yielding sharper structural details and improved overall fidelity. The fine-tuned SPIRiT-based CoilDrop-MRI achieves the best performance, reaching a PSNR of 32.65 dB (Table 2) and outperforming SPIRiT-based SSDU by 0.95 dB. Under the SENSE formulation, CoilDrop-MRI and SSDU achieve comparable PSNR values (~32.0 dB), but CoilDrop-MRI consistently produces visually sharper reconstructions. Overall, the fine-tuned CoilDrop-MRI yields reconstructions that most closely match the BM4D-denoised fully sampled reference (Fig. 8a).

**4.5 Diffusion MRI applicability**

CoilDrop-MRI's efficacy is further validated on dMRI reconstruction tasks. The conventional single-shot Tikhonov-regularized reconstruction under 4-fold acceleration struggles with residual aliasing artifacts and noise amplification; both issues are effectively mitigated in CoilDrop-MRI reconstructions for diffusion-weighted images (DWIs) (Fig. 9). The improved image quality translates to more accurate DTI parameter estimation (Fig. 10), with reduced aliasing in MD and AD maps, and improved delineation of white matter structures in FA maps, particularly near the corpus callosum. CoilDrop-MRI's model-based implementation also allows flexible integration of phase error correction in multi-shot joint reconstruction (Supplementary Fig. S2). By effectively resolving inter-shot phase inconsistencies, 4-shot joint CoilDrop-MRI produces high-quality diffusion images and DTI parameter maps, marginally outperforming the conventional multi-shot Tikhonov-regularized joint reconstruction (Supplementary Fig. S7).

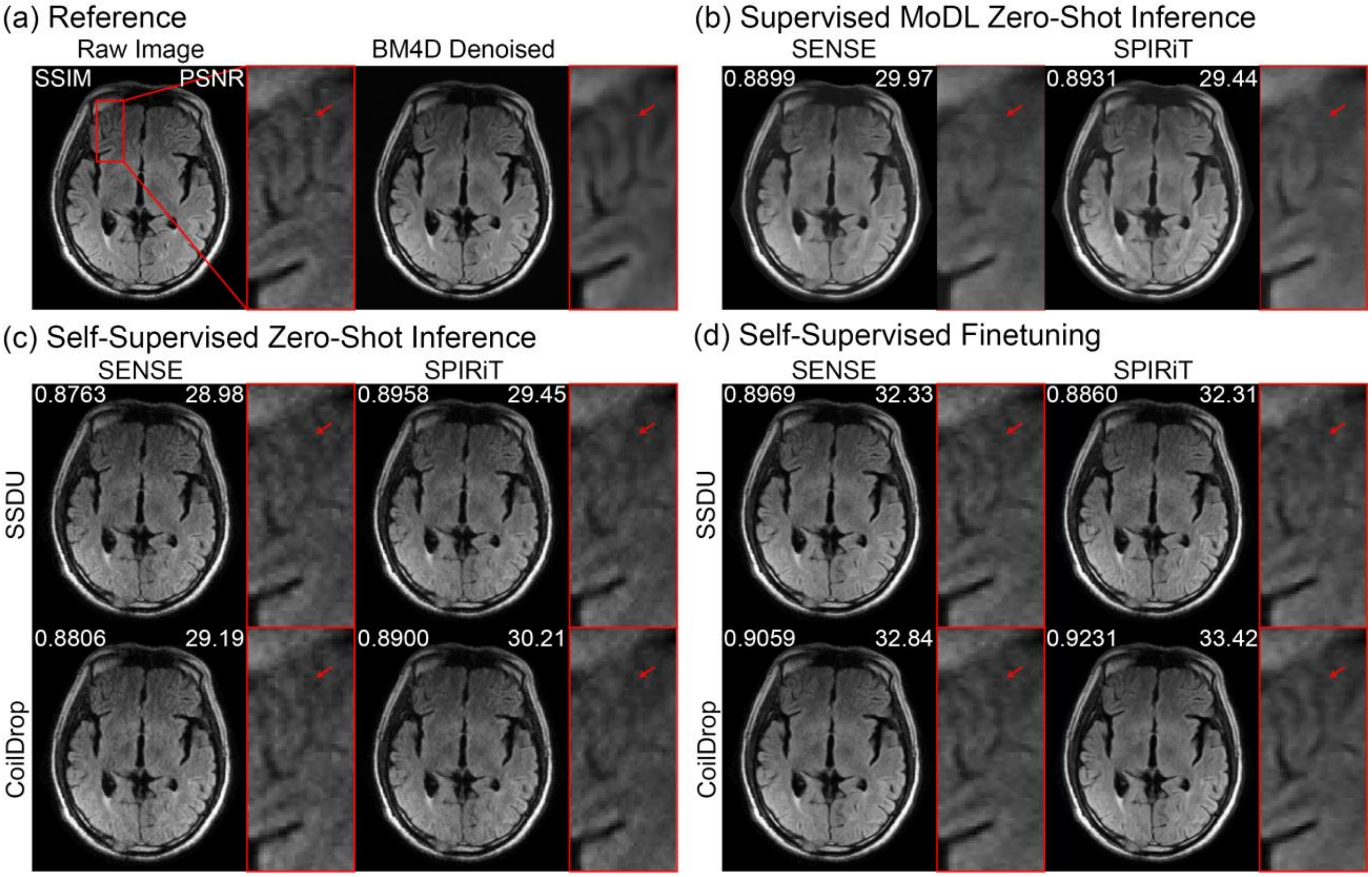

**Figure 8. CoilDrop-MRI generalization to out-of-distribution UCSF 0.55T T2-FLAIR data.** Representative reconstructed images and zoomed-in regions from one test subject are shown under various reconstruction methods: (a) the fully sampled raw image and the BM4D-denoised high-SNR image used as the reference; (b) zero-shot inference using supervised MoDL pre-trained on M4Raw; (c) zero-shot inference using self-supervised SSDU and CoilDrop-MRI pre-trained on M4Raw; and (d) self-supervised fine-tuning on the target 0.55T data using SSDU and CoilDrop-MRI. Both SENSE- and SPIRiT-based formulations are compared. Red arrows highlight a specific anatomical detail to facilitate visual comparison of reconstruction fidelity across methods.

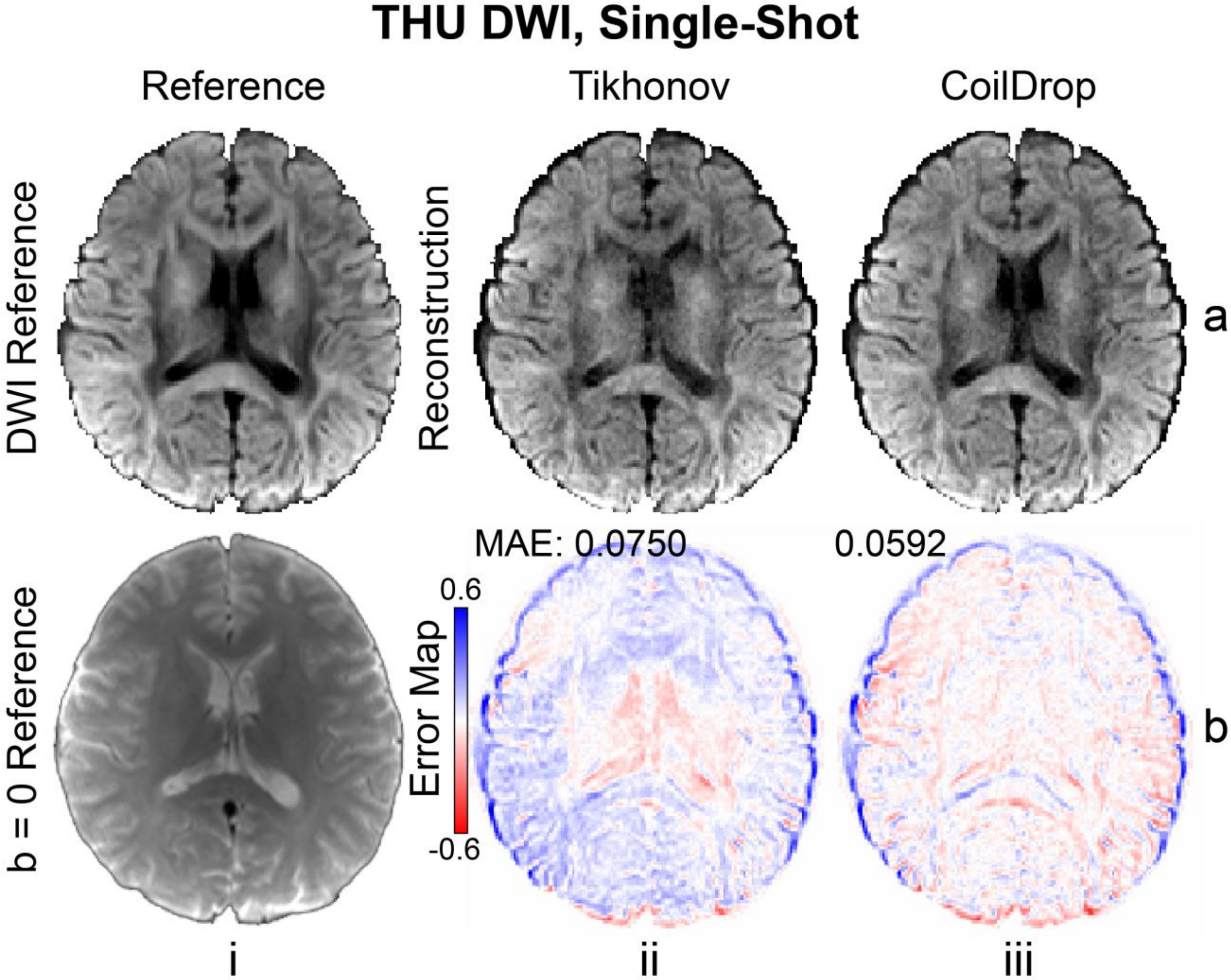

**Figure 9. Comparison of diffusion-weighted image (DWI) reconstructions.** Representative DWI slices reconstructed from single-shot EPI data are shown for comparison. (i) The reference column shows the fully-sampled, multi-repetition-averaged DWI reference at the top; a corresponding b = 0 image is also included at the bottom. The evaluated reconstruction methods include (ii) the Tikhonov-regularized SENSE reconstruction and (iii) the proposed CoilDrop-MRI implemented in the SENSE formulation. Error maps are shown in panels (b-ii) and (b-iii), with the mean absolute error (MAE) annotated in the top-left corner.

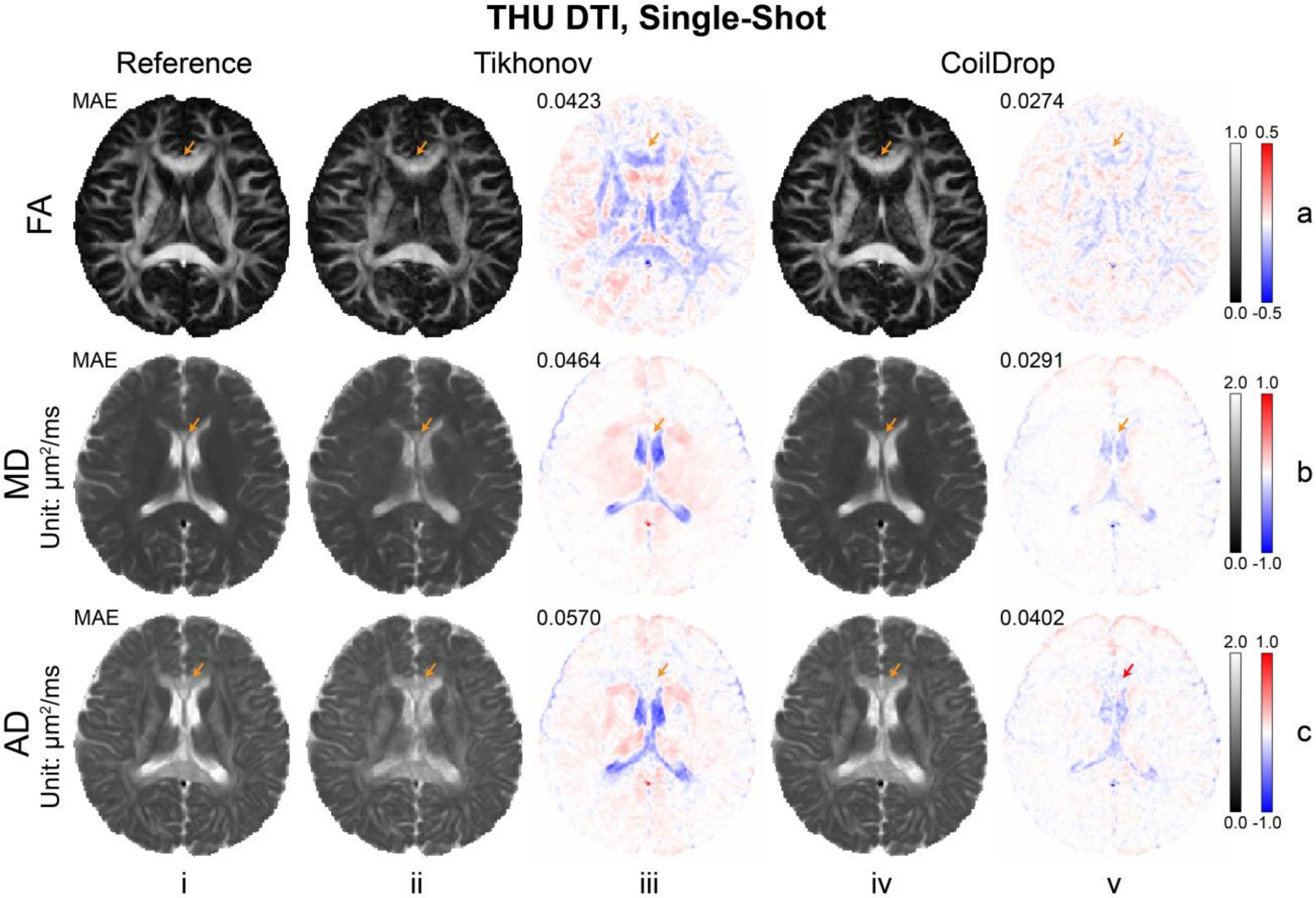

**Figure 10. Comparison of diffusion tensor imaging (DTI) metrics for single-shot diffusion MRI data under 4× acceleration.** Representative fractional anisotropy (FA, a), mean diffusivity (MD, b), and axial diffusivity (AD, c) maps reconstructed using only the first shot from a 4-shot EPI acquisition are compared. The three-repetition-averaged Tikhonov-regularized MUSE reconstruction (i) serves as the high-SNR reference. The evaluated methods include the Tikhonov-regularized SENSE reconstruction (ii) and the proposed CoilDrop-MRI (in the SENSE formulation) reconstruction (iv). The corresponding error maps (iii, v), with the mean absolute error (MAE) values annotated in the top-left corner of each map, are also provided to quantify the reconstruction accuracy. Orange arrows highlight specific anatomical details that are better recovered by CoilDrop-MRI.

## 5. Discussion

In this study, we introduce a new perspective for model-based deep learning MRI reconstruction by explicitly leveraging the coil dimension for self-supervised reconstruction. Building on SSDU, which primarily exploits k-space partitioning, we propose CI-SSDU and CoilDrop-MRI, which both explicitly exploit the coil dimension to train the reconstruction model, and integrate them into both image-domain (SENSE) and k-space (SPIRiT) deep learning reconstruction frameworks. Extensive evaluations across multiple field strengths (0.3T, 0.55T, and 3T) and imaging modalities (T1w, T2w, T2-FLAIR, and dMRI) demonstrate consistent improvements over conventional Tikhonov-regularized reconstruction and existing SSDU-based methods. In particular, CoilDrop-MRI consistently outperforms CI-SSDU and achieves performance approaching that of fully supervised MoDL. Additionally, CoilDrop-MRI is highly data-efficient, requiring only a small number of training subjects, while also demonstrating robust generalization across modalities and scanners. Furthermore, its flexibility in adapting the forward model enables effective handling of more complex scenarios, such as phase-corrected multi-shot dMRI reconstruction. These results highlight CoilDrop-MRI as a promising and versatile approach for advancing parallel MRI reconstruction.

The performance gains of CoilDrop-MRI can be primarily attributed to an improved self-supervised strategy. Fundamentally, the effectiveness of this approach depends largely on the extent to which the partitioning scheme exploits the information redundancy present in the multi-coil k-space data. SSDU, which applies an identical k-space mask across all coils, does not explicitly leverage this inter-coil redundancy. CI-SSDU instead jointly partitions in the k-space and coil dimension, thereby exploiting the rich signal correlations among receiver channels and enabling more effective self-supervised training. CoilDrop performs the partition exclusively in the coil dimension and achieves the best performance in our experiments. One possible explanation is that inter-coil signal correlations are physically grounded in shared anatomy and coil sensitivity profiles, whereas local k-space correlations may be less consistently reliable. More broadly, this perspective suggests that alternative self-supervised reconstruction and denoising strategies could be developed by exploiting other dimensions when available, such as repetitions (Lehtinen et al., 2018; Zihan Li et al., 2026), diffusion encodings (Fadnavis et al., 2020; Mani et al., 2021; Tian et al., 2022; Wu et al., 2019), and the temporal dimension in dynamic imaging (Kang et al., 2024; Tsao et al., 2003; Vizioli et al., 2021).

We implemented CoilDrop-MRI within both SENSE- and SPIRiT-based reconstruction frameworks to leverage the complementary advantages of image-domain and k-space formulations. To date, most model-based deep learning approaches have been built upon SENSE (Aggarwal et al., 2019; Hammernik et al., 2017; Millard & Chiew, 2023, 2024; Yaman et al., 2020), which is computationally efficient but relies on accurate estimation of coil sensitivity maps. In contrast, SPIRiT avoids explicit sensitivity estimation and is therefore more robust in scenarios where such estimation can be challenging, such as low-SNR acquisitions or motion-corrupted data (Hamilton et al., 2017). Additionally, SPIRiT benefits from better conditioning of the inverse problem in k-space, which often leads to reduced g-factor-related noise amplification (Lustig & Pauly, 2010). Consistent with these advantages, SPIRiT-based formulations consistently achieve slightly better quantitative performance and improved visual quality compared to SENSE-based formulations in our experiments (Figs. 3-6, 8; Table 2). Importantly, both formulations retain a model-based structure, enabling flexible incorporation of forward models and regularization terms across diverse acquisition settings.

CoilDrop-MRI further demonstrates strong data efficiency and generalization, both of which benefit from the underlying model-based formulation. By constraining the network with the forward model, the model-based formulation guides learning toward the underlying acquisition physics rather than contrast-specific image features, reducing the reliance on large training datasets. As a result, CoilDrop-MRI achieves high reconstruction quality with limited data, surpassing the Tikhonov baseline when trained on fewer than 5 subjects (Fig. 7). For applications such as dMRI and functional MRI (fMRI), where multiple volumes are available per subject, CoilDrop-MRI can enable effective subject-specific training. In our dMRI experiments, satisfactory performance is achieved by CoilDrop-MRI trained on a single subject in a fully self-supervised manner. Additionally, CoilDrop-MRI exhibits strong generalization across imaging conditions. It generalizes well across contrasts (T1w, T2w, T2-FLAIR), where mixed-modality training yields robust performance (Table 1), and across scanners (0.3T data from M4Raw and 0.55T data acquired at UCSF), where zero-shot inference remains competitive and can be further improved through self-supervised fine-tuning (Fig. 8; Table 2). These properties highlight the practical applicability of CoilDrop-MRI, supporting reliable deployment in diverse clinical and research settings.

The effective joint denoising and reconstruction capability of CoilDrop-MRI makes it particularly well-suited for low-SNR imaging regimes. This is demonstrated in both low-field MRI (0.3T and 0.55T), where SNR is intrinsically limited by the magnetic field strength especially under high acceleration factors (R = 2 and 3 with only 4 receive channels), and dMRI, where signal is substantially attenuated due to diffusion encoding. In these challenging settings, CoilDrop-MRI consistently achieves high-fidelity reconstructions, highlighting its robustness to noise-dominated conditions. These results suggest that CoilDrop-MRI is applicable to other SNR-limited applications, such as high-resolution, submillimeter dMRI (Dong et al., 2025; Ziyu Li et al., 2026; Liao et al., 2023) and fMRI (Bandettini et al., 2021; Kay et al., 2019; Vizioli et al., 2021), where maintaining image quality as voxel size decreases remains challenging. Furthermore, recent advances in MRI receiver coil design (Frass-Kriegl et al., 2018; Scholz et al., 2021; Wu et al., 2024) that optimize inter-coil signal and noise characteristics may further enhance the benefits of CoilDrop-MRI, pointing to promising opportunities at the intersection of hardware and reconstruction.

## 6. Conclusion

In this study, we introduce CoilDrop-MRI, a coil partitioning-based self-supervised reconstruction framework for model-based deep learning MRI reconstruction. Building on SSDU, we extend self-supervised learning to the coil dimension and integrate the approach into both image-domain (SENSE-based) and k-space (SPIRiT-based) frameworks. Across multiple field strengths and imaging contrasts, CoilDrop-MRI consistently outperforms conventional Tikhonov and state-of-the-art SSDU methods, and approaches the performance of fully supervised MoDL. It further demonstrates strong data efficiency, robust cross-modality and cross-scanner generalization, and flexibility in handling complex forward models such as multi-shot dMRI. In conclusion, CoilDrop-MRI provides a robust, practical, and versatile framework for advancing parallel MRI reconstruction.

**Acknowledgments**

This work was supported by the National Natural Science Foundation of China (Grant No. 82302166), Beijing Medical and Health Science and Technology Promotion Center (Grant No. 2024-3-5031), Tsinghua University Startup Fund, Tsinghua University Dushi Program (Grant No. 20241080026, 20251080056), and the Wellcome Trust (Grant No. WT327832/Z/25/Z). Initial results of this research were submitted on October 29, 2025, to the annual scientific meeting of the International Society of Magnetic Resonance in Medicine (ISMRM) and were accepted for an oral presentation in May 2026.

**Code availability**

The source code will be made publicly available upon revision.

**Declaration of generative AI**

During the preparation of this manuscript, the authors used ChatGPT for the purpose of language polishing and improving readability. The tool was not used for generating scientific content, analyzing data, or drawing conclusions. The authors reviewed and edited all AI-assisted output and take full responsibility for the content of this publication.

## Supplementary information

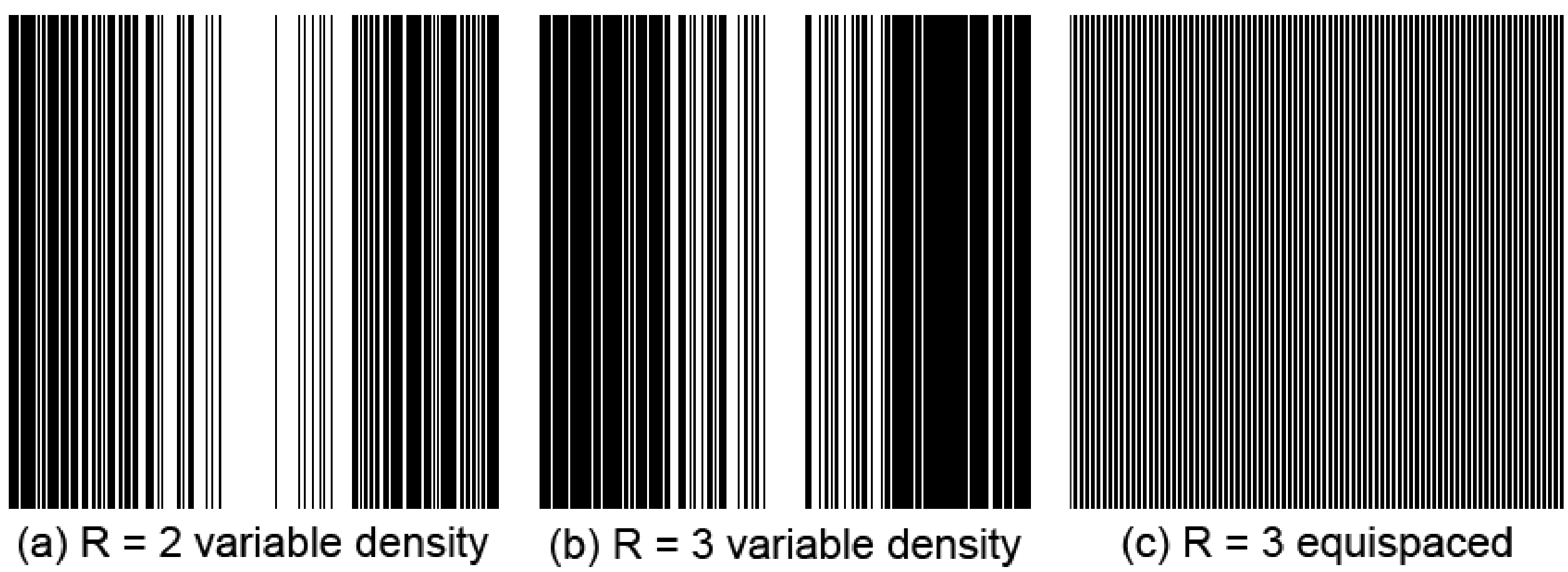

**Supplementary Figure S1. Retrospective k-space undersampling masks.** Sampling patterns applied along the phase-encoding direction for: (a) variable density (VD) sampling at 2× acceleration (R = 2); (b) VD sampling at 3× acceleration (R = 3); and (c) equispaced sampling at 3× acceleration (R = 3). White lines denote sampled phase-encoding positions.

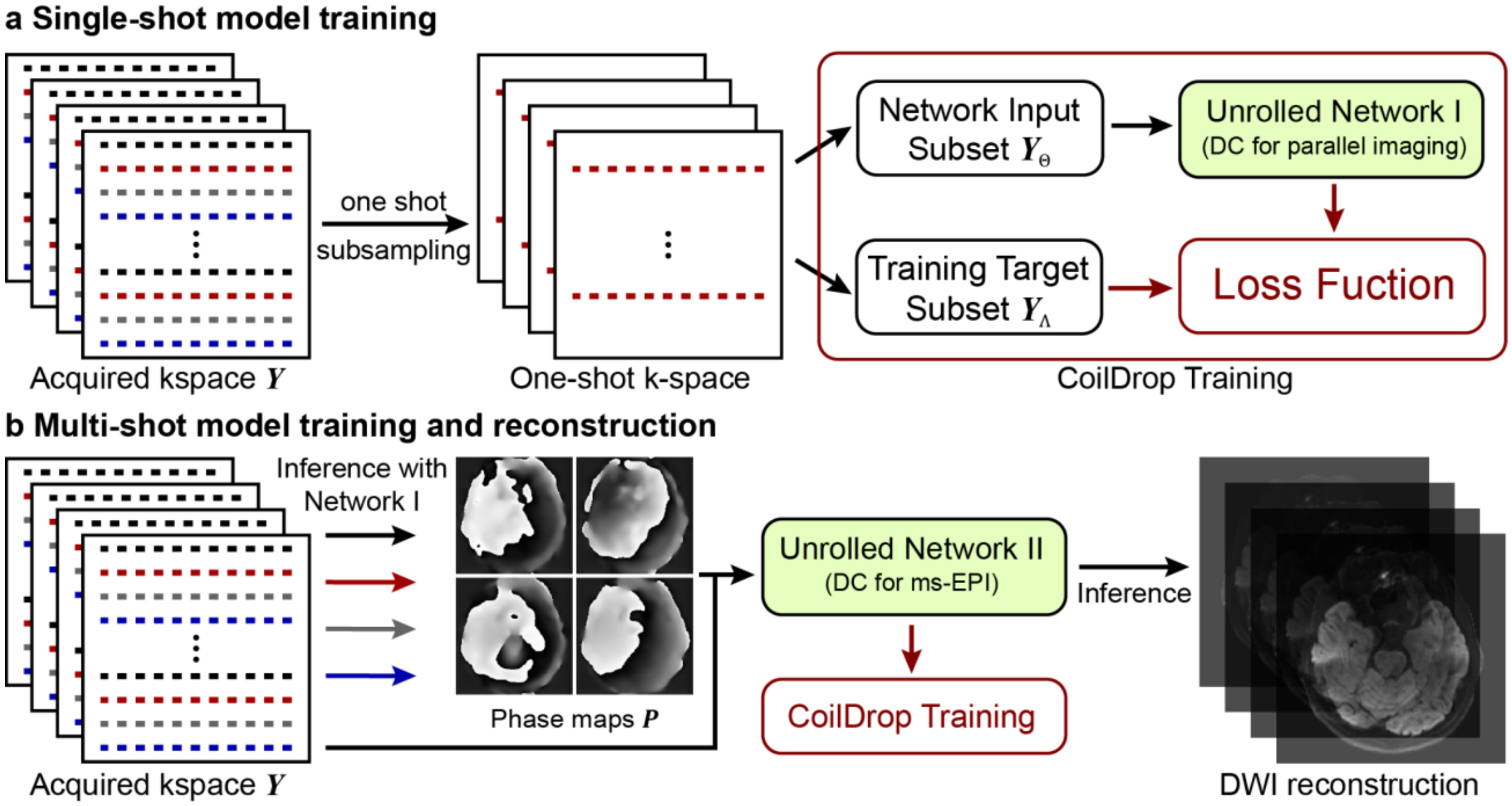

**Supplementary Figure S2. Two-stage CoilDrop-MRI reconstruction framework for multi-shot diffusion MRI.** (a) Stage 1: single-shot reconstruction model training for shot-phase estimation. The acquired multi-shot k-space data ($\boldsymbol{Y}$) are first separated into individual single-shot data. Each shot is then used to train a standard parallel imaging reconstruction model (Unrolled Network I) via the CoilDrop strategy. (b) Stage 2: multi-shot joint reconstruction training. The trained Unrolled Network I reconstructs each shot separately, enabling the extraction of shot-specific phase maps $\boldsymbol{P}$. These phase maps, together with the multi-shot k-space data, are fed into a phase-corrected multi-shot reconstruction model (Unrolled Network II) for CoilDrop training and inference.

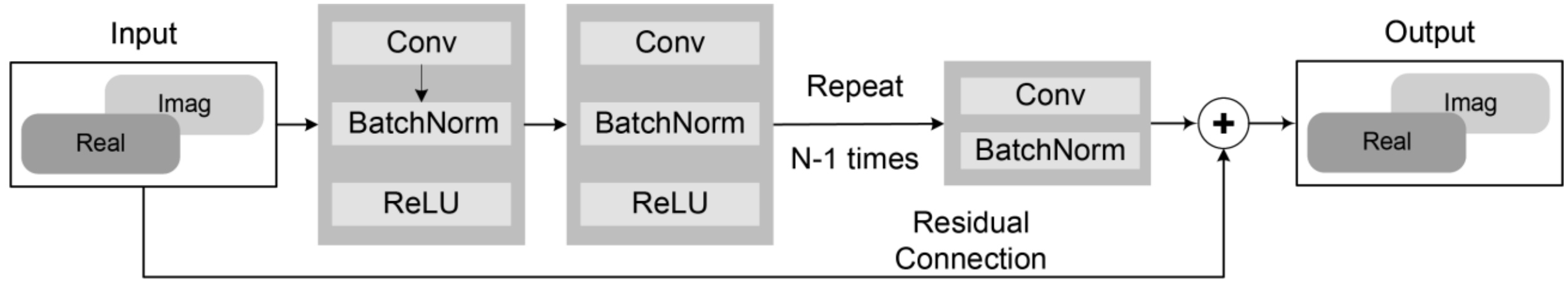

**Supplementary Figure S3. Architecture of the CNN denoiser.** Complex-valued MR images are represented as two-channel inputs by separating the real and imaginary components. The denoiser consists of a $N$-layer CNN: the first $N-1$ layers each contain convolution, batch normalization, and ReLU activation, while the final layer contains convolution and batch normalization only, without ReLU. A residual connection adds the input to the network output to produce the final output complex-valued image.

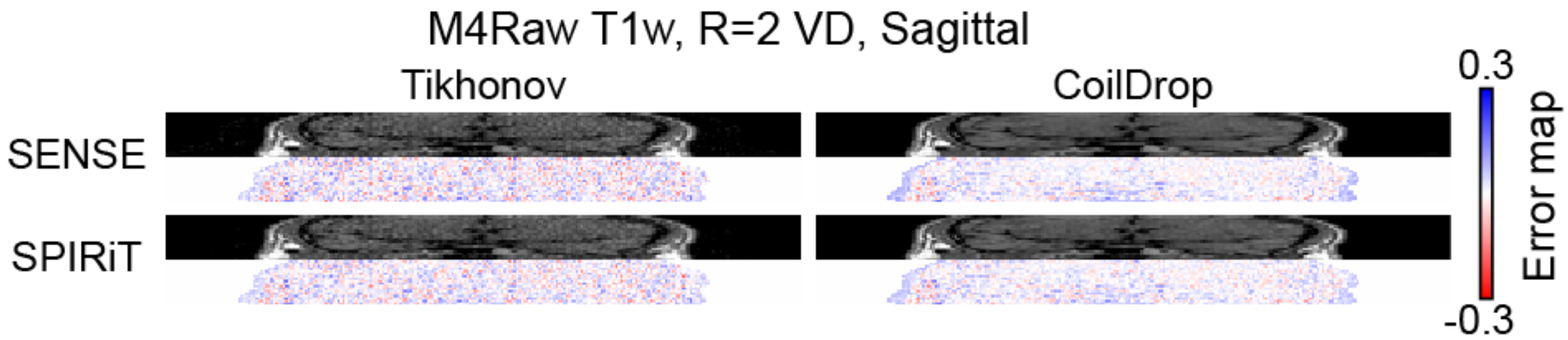

**Supplementary Figure S4. Sagittal reconstructions from M4Raw T1w data with 2-fold variable-density (VD) undersampling.** Tikhonov-regularized reconstruction (left) and the proposed CoilDrop-MRI (right) are compared under both the SENSE and SPIRiT formulations. Error maps with respect to the fully sampled reference are displayed below each reconstruction.

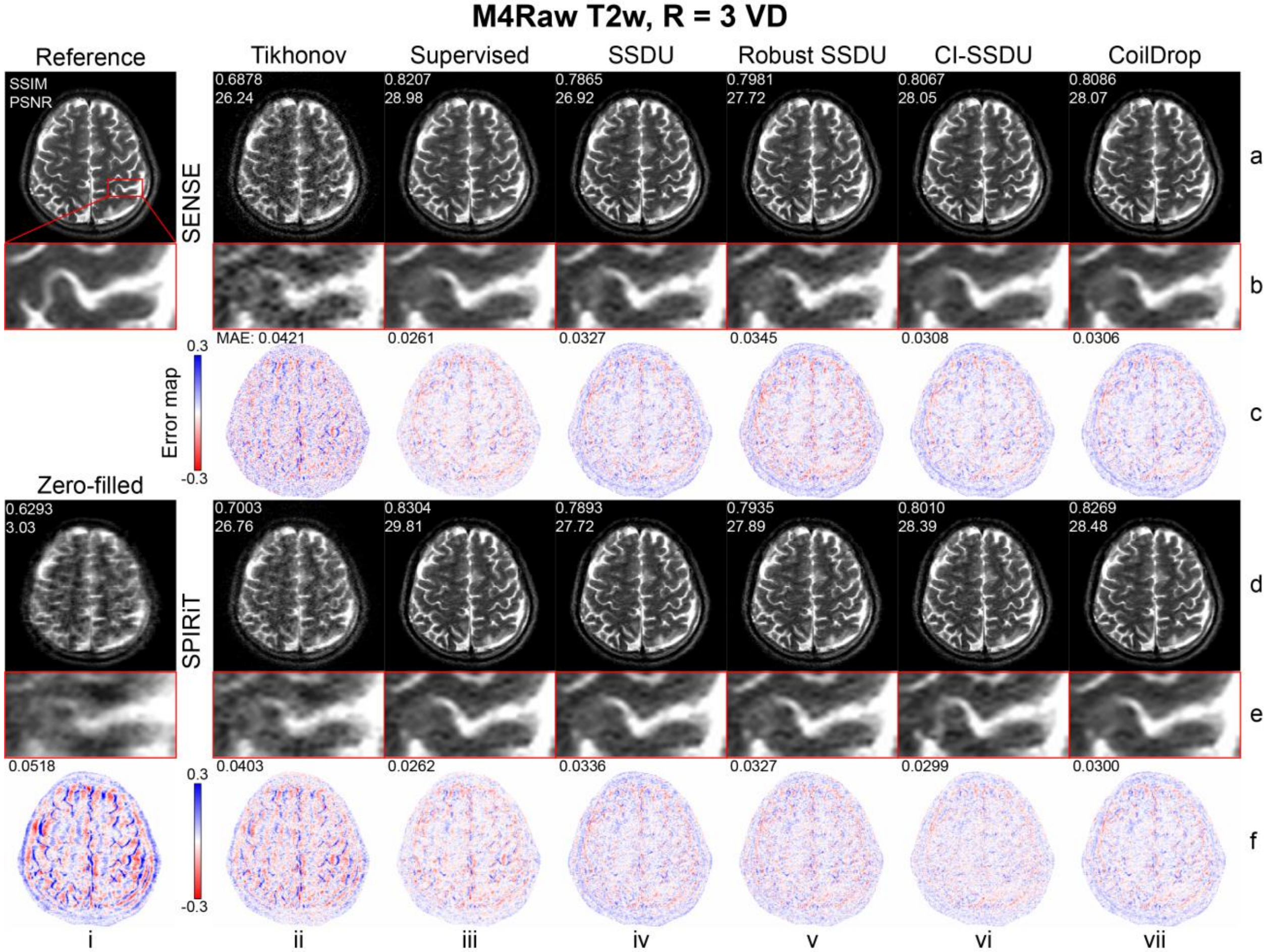

**Supplementary Figure S5. Comparison of reconstruction methods with variable density (VD) sampling at 3× acceleration (R = 3) for T2-weighted (T2w) images.** Representative reconstructed images, zoomed-in regions, and corresponding error maps are shown for R = 3 VD T2w images from one test subject. Both SENSE- and SPIRiT-based implementations are compared for the following: multi-repetition-averaged high-SNR reference (top left), zero-filled reconstruction (bottom left), Tikhonov-regularized reconstruction, supervised MoDL, and four self-supervised strategies: SSDU, Robust SSDU, Coil-Incoherent SSDU (CI-SSDU), and CoilDrop. Quantitative metrics, including structural similarity index (SSIM), peak signal-to-noise ratio (PSNR), and mean absolute error (MAE) with the high-SNR reference are calculated to quantify reconstruction fidelity.

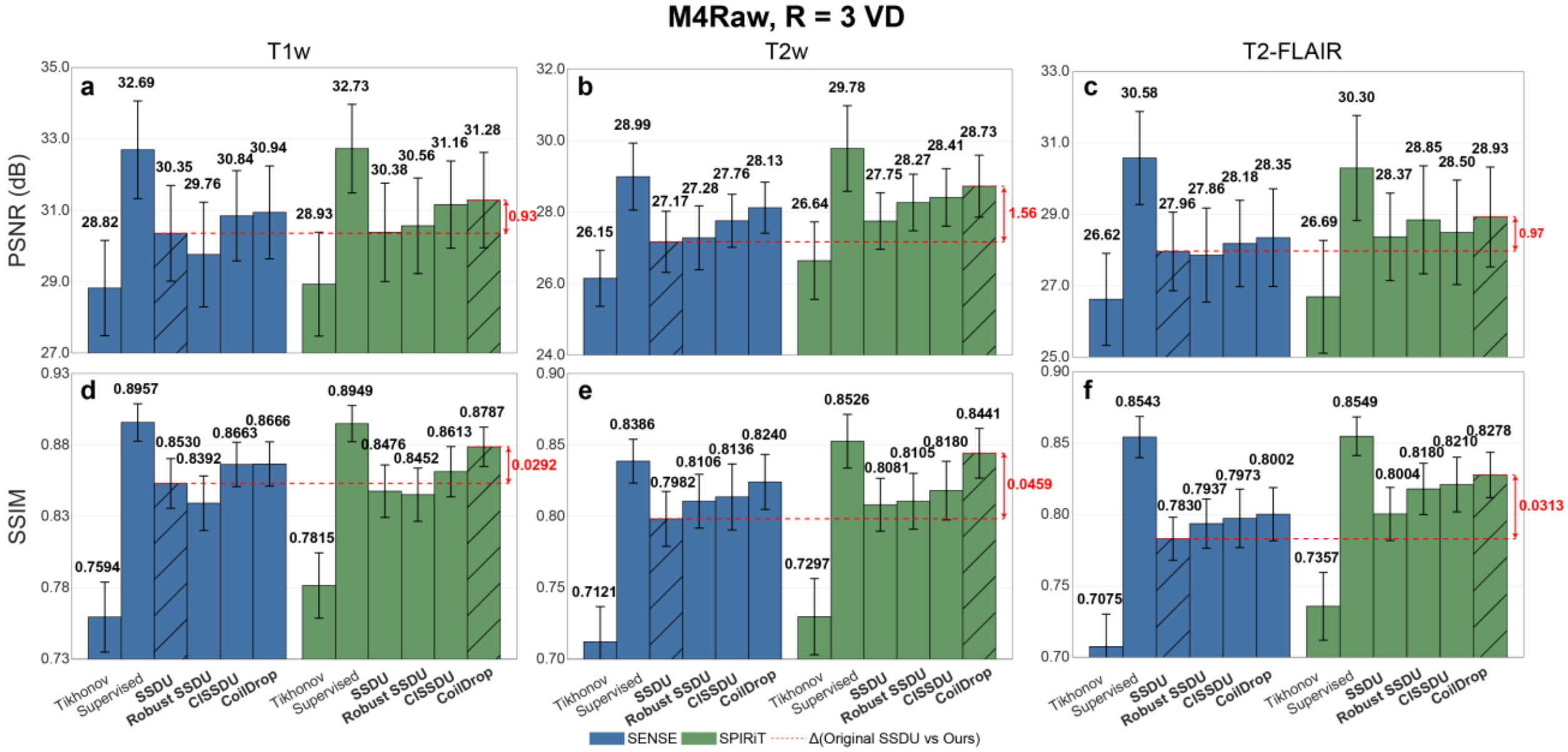

**Supplementary Figure S6. Quantitative comparison of reconstruction methods under variable density (VD) sampling at 3× acceleration (R = 3).** Bar charts illustrate the average (a-c, top row) peak signal-to-noise ratio (PSNR) and (d-f, bottom row) structural similarity index (SSIM) across the test subjects for (a, d) T1-weighted (T1w), (b, e) T2-weighted (T2w), and (c, f) T2-FLAIR images under R = 3 VD reconstructions. Blue and green bars represent reconstruction frameworks employing SENSE- and SPIRiT-based data consistency, respectively. The evaluated methods include the conventional Tikhonov-regularized baseline, the supervised MoDL, and four self-supervised strategies (marked in **bold**): SSDU, Robust SSDU, Coil-Incoherent SSDU (CI-SSDU), and the proposed CoilDrop (highlighted with diagonal hatches). Red dashed lines and text indicate the quantitative performance gains of the proposed CoilDrop-MRI over the original SSDU method. Error bars denote standard deviations.

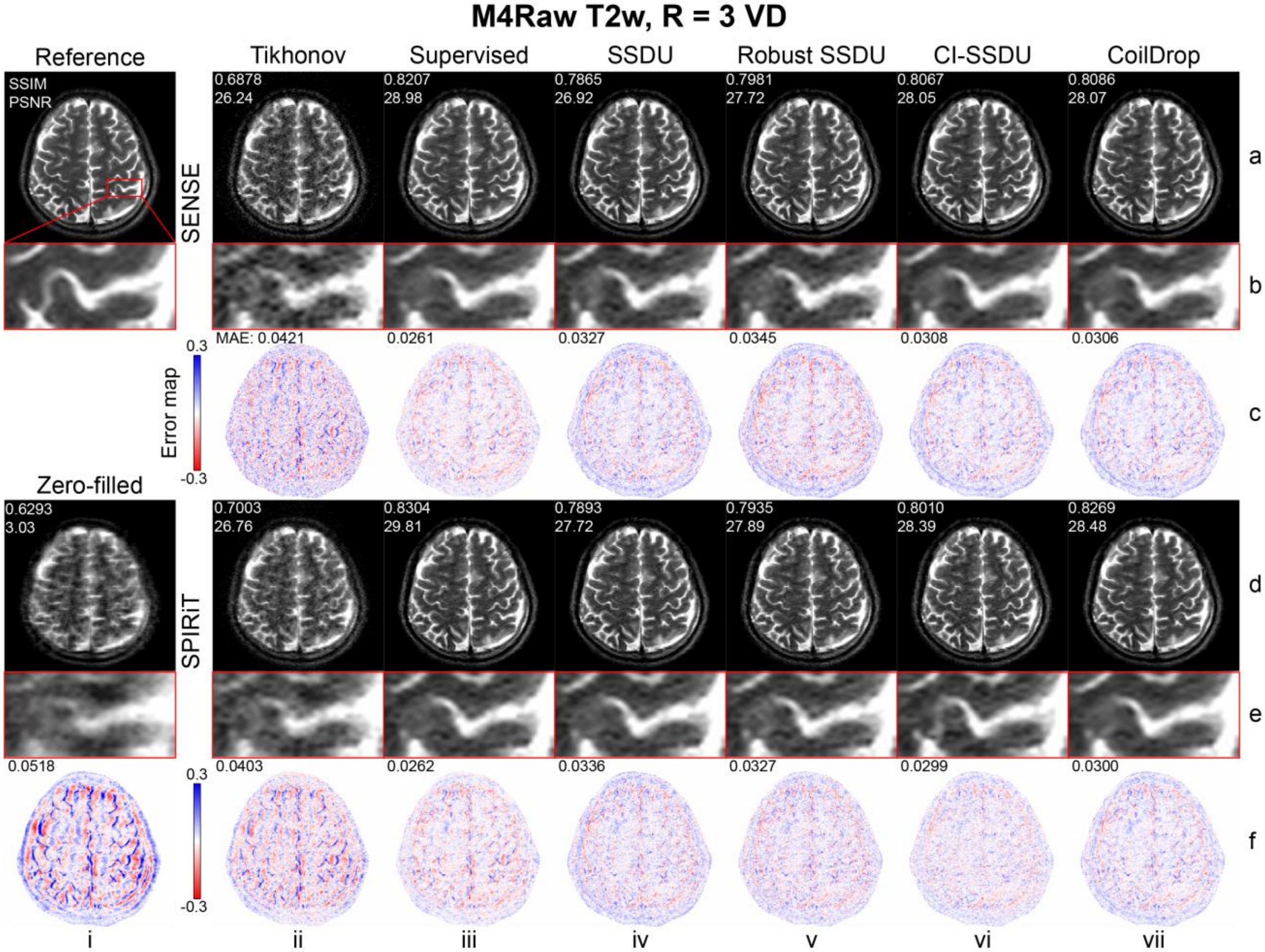

**Supplementary Figure S7. Comparison of diffusion tensor imaging (DTI) metrics for multi-shot diffusion MRI data.** Representative fractional anisotropy (FA, a), mean diffusivity (MD, b), and axial diffusivity (AD, c) maps reconstructed from 4-shot echo-planar imaging (EPI) data are compared. The three-repetition-averaged Tikhonov-regularized MUSE reconstruction (i) serves as the high-SNR reference. The evaluated methods include the Tikhonov-regularized MUSE reconstruction (ii) and the proposed multi-shot CoilDrop-MRI reconstruction (iv). The corresponding error maps (iii, v), with the mean absolute error (MAE) values annotated in the top-left corner of each map, are also provided to quantify the reconstruction accuracy. Orange arrows highlight specific anatomical details that are better recovered by CoilDrop-MRI.